\documentclass[manuscript,screen]{acmart}

\AtBeginDocument{%
  \providecommand\BibTeX{{%
    \normalfont B\kern-0.5em{\scshape i\kern-0.25em b}\kern-0.8em\TeX}}}

\setcopyright{acmcopyright}
\copyrightyear{2018}
\acmYear{2018}
\acmDOI{XXXXXXX.XXXXXXX}

\usepackage{hyperref}
\usepackage{multirow}

\usepackage{graphicx}
\usepackage{xcolor,amsmath}
\usepackage[linesnumbered,ruled,vlined]{algorithm2e}
\DontPrintSemicolon

\renewcommand{\KwSty}[1]{\textnormal{\textcolor{blue!90!black}{\ttfamily\bfseries #1}}\unskip}

\SetKwComment{Comment}{\color{green!50!black}\# }{}
\renewcommand{\CommentSty}[1]{\textnormal{\ttfamily\color{green!50!black}#1}\unskip}

\newcommand{\var}{\texttt}
\newcommand{\FuncCall}[2]{\texttt{\bfseries #1(#2)}}
\SetKwProg{Function}{def}{:}{}
\renewcommand{\ProgSty}[1]{\texttt{\bfseries #1}}
\SetKwProg{For}{for}{:}{}
\SetKwProg{If}{if}{:}{}

\newtheorem{example}{Example}

\usepackage{url}
\newcommand{\TheName}{\textbf{LCaC}}

\begin{document}

\title{Natural Language based Context Modeling and Reasoning for Ubiquitous Computing with Large Language Models: A Tutorial}

\author{Haoyi Xiong}
\email{haoyi.xiong.fr@ieee.org}
\orcid{0000-0002-5451-3253}
\author{Jiang Bian}
\email{jiangbian03@gmail.com}
\affiliation{%
  \institution{Big Data Lab, Baidu, Inc.}
  \streetaddress{No.10 Xibeiwang East Road}
  \city{Haidian}
  \state{Beijing}
  \country{China}
  \postcode{100085}
}

\author{Sijia Yang}
\email{anneyang@bupt.edu.cn}
\affiliation{%
  \institution{Beijing University of Posts and Telecommunications}
  \streetaddress{10 Xitucheng Rd}
  \city{Haidian}
  \state{Beijing}
  \country{China}
  \postcode{100876}
}

\author{Xiaofei Zhang}
\email{zhangxiaofei01@baidu.com}
\affiliation{%
  \institution{Baidu, Inc.}
  \streetaddress{No.10 Xibeiwang East Road}
  \city{Haidian}
  \state{Beijing}
  \country{China}
  \postcode{100085}
}

\author{Linghe Kong}
\email{linghe.kong@sjtu.edu.cn}
\affiliation{%
  \institution{Department of Computer Science and Engineering, Shanghai Jiaotong University}
  \streetaddress{800 Dongchuan Road}
  \city{Minghang}
  \state{Shanghai}
  \country{China}
  \postcode{200240}
}

\author{Daqing Zhang}
\email{daqing.zhang@telecom-sudparis.eu}
\affiliation{%
  \institution{T{\'e}l{\'e}com SudParis, Institut Polytechnique de Paris}
  \streetaddress{9 Rue Charles Fourier}
  \city{Evry Courcouronnes}
  \state{Essonne}
  \country{France}
  \postcode{91000}
}
\renewcommand{\shortauthors}{Xiong et al.}

\begin{abstract}
Large language models (LLMs) have become phenomenally surging\footnote{https://huggingface.co/blog/large-language-models}, since 2018 -- two decades after introducing context-awareness into computing systems~\cite{dey1998cyberdesk,abowd1998context,abowd1999towards,dey2001conceptual}. Through taking into account the situations of ubiquitous devices, users and the societies, context-aware computing has enabled a wide spectrum of innovative applications, such as assisted living, location-based social network services and so on. To recognize contexts and make decisions for actions accordingly, various artificial intelligence technologies, such as Ontology and OWL, have been adopted as representations for context modeling and reasoning~\cite{wang2004ontology,christopoulou2005ontology}. Recently, with the rise of LLMs and their improved natural language understanding and reasoning capabilities, it has become feasible to model contexts using natural language and perform context reasoning by interacting with LLMs such as ChatGPT and GPT-4. 

In this tutorial, we demonstrate the use of texts, prompts, and autonomous agents (AutoAgents) that enable LLMs to perform context modeling and reasoning without requiring fine-tuning of the model. We organize and introduce works in the related field, and name this computing paradigm as the \underline{L}LM-driven \underline{C}ontext-\underline{a}ware \underline{C}omputing (\TheName{}). In the \TheName{} paradigm, users' requests, sensors reading data, and the command to actuators are supposed to be represented as texts. Given the text of users' request and sensor data, the AutoAgent models the context by prompting and sends to the LLM for context reasoning. LLM generates a plan of actions and responds to the AutoAgent, which later follows the action plan to foster context-awareness. To prove the concepts, we use two showcases -- (1) operating a mobile z-arm in an apartment for assisted living, and (2) planning a trip and scheduling the itinerary in a context-aware and personalized manner. Furthermore, we analyze several factors that might affect the performance of LLM-driven context-awareness, and then discuss the future research directions.
\end{abstract}




\maketitle

%Perceptual models, such as BERT, particularly excel in text comprehension due to their bidirectional attention flows~\cite{kenton2019bert}. In contrast, Generative Pre-trained Transformer (GPT) series, which includes models such as ChatGPT and GPT-4 ~\cite{koziolek2023chatgpt,openai2023gpt4}, excel in text generation through the unidirectional attentions. These models have potential applications in question-answering and chat services ~\cite{radford2018improving,openai2023gpt4}. Based on GPT, one can try to describe the \emph{problem to solve} in texts sent to the model and obtain a \emph{solution to the problem} from the LLM's response. The model might be able to respond the question correctly based on its capability of reasoning and knowledge obtained in pre-training. This phenomena is called \emph{emergence} and \emph{homogenization} -- the model manifests itself instead of being deliberately programmed while unanticipated properties may emerge~\cite{bommasani2021opportunities}. Furthermore, to make LLMs adapt to new tasks unseen in the training datasets, a new problem-solving paradigm, named few-shot prompting~\cite{reynolds2021prompt,schick2022true}, has been invented; one could attach similar problems and their solutions as ``prompts'' with the description of the original problem, to help LLMs learn from examples without requiring further fine-tuning of the model. 

\section{Introduction}
In recent years, particularly since 2018, \underline{L}arge \underline{L}anguage \underline{M}odels (LLMs) or foundation models, from ELMO, BERT~\cite{kenton2019bert}, GPT-2 and 3, Turing-NLG, to ChatGPT, ERNIE Bot, GPT-4~\cite{openai2023gpt4} and the upcoming GPT-5, have experienced a phenomenal surge in both popularity and advancements. These cutting-edge LLMs are revolutionizing the field of natural language processing (NLP), enabling machines to understand, interpret, and generate human-like texts. Basically two types of LLMs have been studied.
%
%While these LLMs are built on top of transformer architectures with attention mechanisms equipped, they could be categorized into two types, perception and generation, with respect to their purposes. 
%
Perceptual models like BERT excel at text comprehension via bidirectional attention flows~\cite{kenton2019bert}, while the GPT series, including ChatGPT and GPT-4, standout in text generation utilizing unidirectional attentions~\cite{koziolek2023chatgpt,openai2023gpt4}. An intriguing quality of GPT-based models, referred to as \emph{emergence} and \emph{homogenization}, is that they can return novel answers to posed problems using internal reasoning and prior knowledge~\cite{bommasani2021opportunities,bubeck2023sparks}. Furthermore, the models can adapt to unfamiliar tasks using few-shot prompting (aka in-context learning), where similar problem and solution pairs are presented as prompts to guide the learning process without the need for model fine-tuning~\cite{reynolds2021prompt,schick2022true}.   

%~\cite{ranganathan2003middleware}
In addition to content understanding and generation, LLMs, more recently, have been studied as a way to implement \emph{autonomous agents} (AutoAgents)~\cite{li2023autonomous} capable of interacting with the environment via physical embodiment~\cite{franklin1997autonomous}. By integrating prompt templates, conversational agents, conversational memory, external knowledgebase and interfaces of tool-uses (e.g., APIs for other AI models, sensors and actuators) within a \emph{LangChain}~\cite{langchain2023}, LLMs extend their capacities, from text comprehension to sensing, perceiving, and acting within a physical environment~\cite{singh2023progprompt,huang2023visual,vemprala2023chatgpt,biggie2023tell}. Moreover, fancy and advanced prompting techniques, such as Iterative Decoding, Chain of thought (CoT), Tree of Thoughts (ToT) and ReAct, further boost LLM-powered agents for task decomposition and planning~\cite{wei2022chain,yao2023tree,yao2022react}. As a result of extensive tool-uses, an LLM-powered AutoAgent can handle a multitude of physical tasks, from reading maps and instructing robotics to supervising vehicular movement and coding for controllers in various industries~\cite{li2023autonomous,singh2023progprompt,huang2023instruct2act,de2023llm,song2022llm,koziolek2023chatgpt,liang2023code}.

%On the other hand, the extensive tool-uses make LLM-powered AutoAgents possible to command a wide range of physical beings, such as reading geographical maps~\cite{li2023autonomous}, generating policies to instruct robotics~\cite{singh2023progprompt,huang2023instruct2act}, overseeing PID controllers for moving vehicles~\cite{de2023llm}, generating the layout of objects for grounded planning~\cite{song2022llm,lin2023grounded}, or generating program codes to command Programmable Logic Controller (PLC) for industry~\cite{koziolek2023chatgpt} and other controllers~\cite{liang2023code}. 

\begin{figure*}
    \centering
    \includegraphics[width=\textwidth]{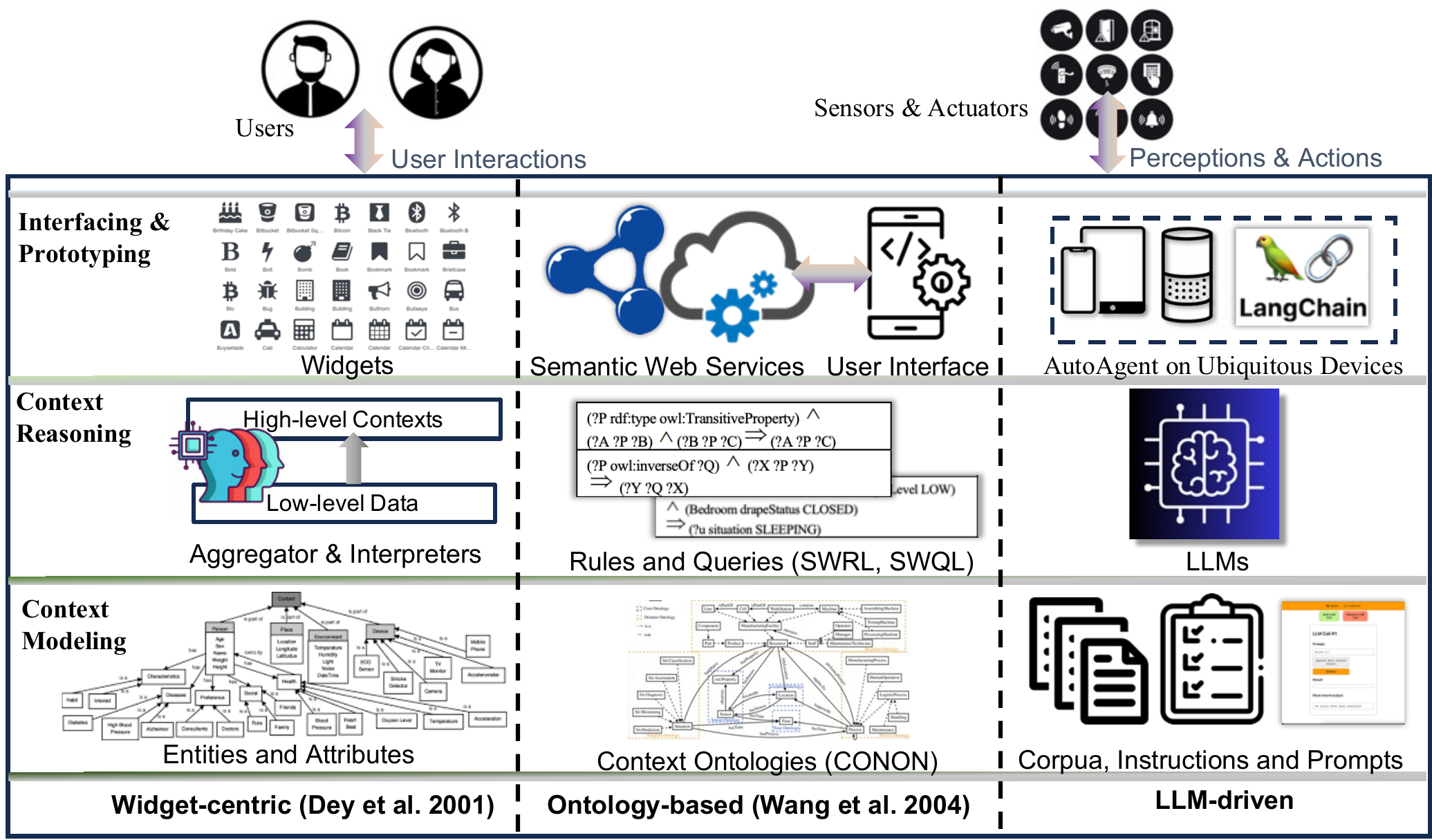}
    \caption{The Three Layers of Conceptual Design in Context-aware Computing and the Mapping of Concepts among Widget-centric~\cite{dey2001conceptual}, Ontology-based~\cite{wang2004ontology} and LLM-driven (Ours) Approaches}
    \label{fig:concept-mapping}
\end{figure*}

Upon above advantages of LLMs and LLM-powered AutoAgents, this work is particularly interested in revisiting \emph{context-awareness}~\cite{abowd1999towards,dey2001conceptual}, a core research area of ubiquitous computing~\cite{abowd2000charting}. Starting from 1998, Dey and Abowd together with other colleagues published a series of seminar work on rapid-prototyping context-aware services or applications and, for the first time, coincided the term \emph{``context-aware computing''} in referred publications~\cite{dey1998cyberdesk,abowd1998context,salber1999context}. In the historical panel at HUC'99~\cite{gellersen1999handheld} (which has since evolved into the ACM UbiComp conference), Abowd and Dey served as panel moderators, where they discussed the definitions and challenges of \emph{context-awareness}~\cite{abowd1999towards} with researchers from ubiquitous computing and computer-human interaction domains. 
Context as referred to in computing terms, signifies the ``implicit situational information'' that covers various information sources, such as physical environment, locations, time, user activities, preferences, and social settings, aiding in the understanding of the environment where computing takes place. Context-awareness indicates the ability of a system to perceive and respond to contextual changes. In essence, a context-aware computing system should be efficient at collecting, analysing, and leveraging context to adapt its services to the user coherently and promptly~\cite{dey2001understanding}.

%a context-aware computing system is built-up with three core components: (1) (close to widgets in Dey's primitives~\cite{dey2001conceptual}),  and (3) . As shown in Fig.~\ref{fig:concept-mapping}, \emph{ Agent}~\cite{ranganathan2003middleware,chen2003intelligent,soldatos2007agent,ziebart2008maximum,lee2009context} and 
%\emph{Ontology}~\cite{chen2003ontology,wang2004ontology,ngo2004developing,christopoulou2005ontology,ejigu2007ontology} technologies have been adopted and practiced to facilitate the implementation of context-aware computing systems.
%
%
%(equivalent to \emph{inference} and \emph{derivation} in~\cite{dey2001conceptual})

Following the conceptual framework~\cite{dey2001conceptual,dey2001understanding} developed by the same group of researchers and best practices in the area~\cite{ranganathan2003middleware,chen2003intelligent,wang2004ontology,christopoulou2005ontology,dey2006icap,ziebart2008maximum}, one can simply categorize the components of context-aware computing systems into three layers: (1) \emph{context modeling} that represents the changing contexts with data for storage, retrieval and computation purposes; (2) \emph{context reasoning} that aggregates and interprets data for generating new understanding on the contexts; and (3) \emph{context inferfacing} that interacts with users, sensors \& actuators for context acquisition and context-enabled actions. Fig.~\ref{fig:concept-mapping} illustrates widget-centric~\cite{dey2001conceptual} and ontology-based~\cite{wang2004ontology} approaches to implement the three layers.
In this work, with respects to the ability of LLMs for language understanding, reasoning and planning, we try to map the LLM-driven technologies into the three layers of context-aware computing, while seeking the feasibility to incorporate natural languages together with LLMs for context modeling and reasoning. The intuitions are quite straight-forward as follows.
\begin{itemize}
    \item \emph{Context Modeling:} As shown in Fig.~\ref{fig:concept-mapping}, to represent contexts for modeling and reasoning purposes, previous studies proposed leveraging either tuples~\cite{dey2001conceptual,dey2001understanding} encompassing entities and attributes or semantic web data/Ontology~\cite{wang2004ontology,christopoulou2005ontology}, such as Web Ontology Language (OWL)~\cite{mcguinness2004owl}. It is reasonable to assume the use of natural language (voice or texts) on top of LLMs could represent contexts despite the context ambiguity issues~\cite{dey2001conceptual}. To effectively model contexts with LLMs, one could pre-train LLMs using large-scale corpus documenting common senses and elements of the world, fine-tune LLMs with instructions for domain-specific contextual information, and invoke LLMs with prompts for adapting the change of contexts.
    
    %represent the contexts using tuples encompassing entities and attributes~\cite{dey2001conceptual,dey2001understanding} or semantic web data, such as Ontology and OWL~\cite{wang2004ontology}, for modeling and reasoning purposes. and it is always not a bad idea to use natural language, such as human-like texts, to represent the contexts for the same purposes. 

    \item \emph{Context Reasoning:} As shown in Fig.~\ref{fig:concept-mapping}, as the proxies of users and sensors \& actuators, Dey et al.~\cite{dey2001conceptual} proposed ``Widgets'' that leveraged rule-based context reasoning with high-level contexts extracted from tuples. Similarly, the ontology-based approach~\cite{wang2004ontology} integrates the user interface and sensors \& actuators via Semantic Web Services, which allows ontology-based reasoning with rules and queries upon standard data formats, such as  and Semantic Web Rule Language (SWRL)~\cite{o2005supporting}.
    %
    %has suggested already \emph{agents} as a way to foster context-awareness in a cyber-physical environment. 
    %
    As the rule-based solutions were frequently lack of generalizability or extensibility, researchers also tried to advance context reasoning with machine learning techniques~\cite{ziebart2008maximum}. However, due to the limited AI power in past several decades, it was difficult to implement effective context reasoner for general-purpose. Fortunately, the sparks of artificial general intelligence (AGI) surged by LLMs~\cite{bubeck2023sparks} make LLM-powered AutoAgent a feasible way of context reasoning. %with nature language.

    %When introducing the concept of context-aware computing, an early work from Dey and Abowd in 2001~\cite{dey2001conceptual} 

\end{itemize}

\begin{figure*}
    \centering
    \includegraphics[width=0.7\textwidth]{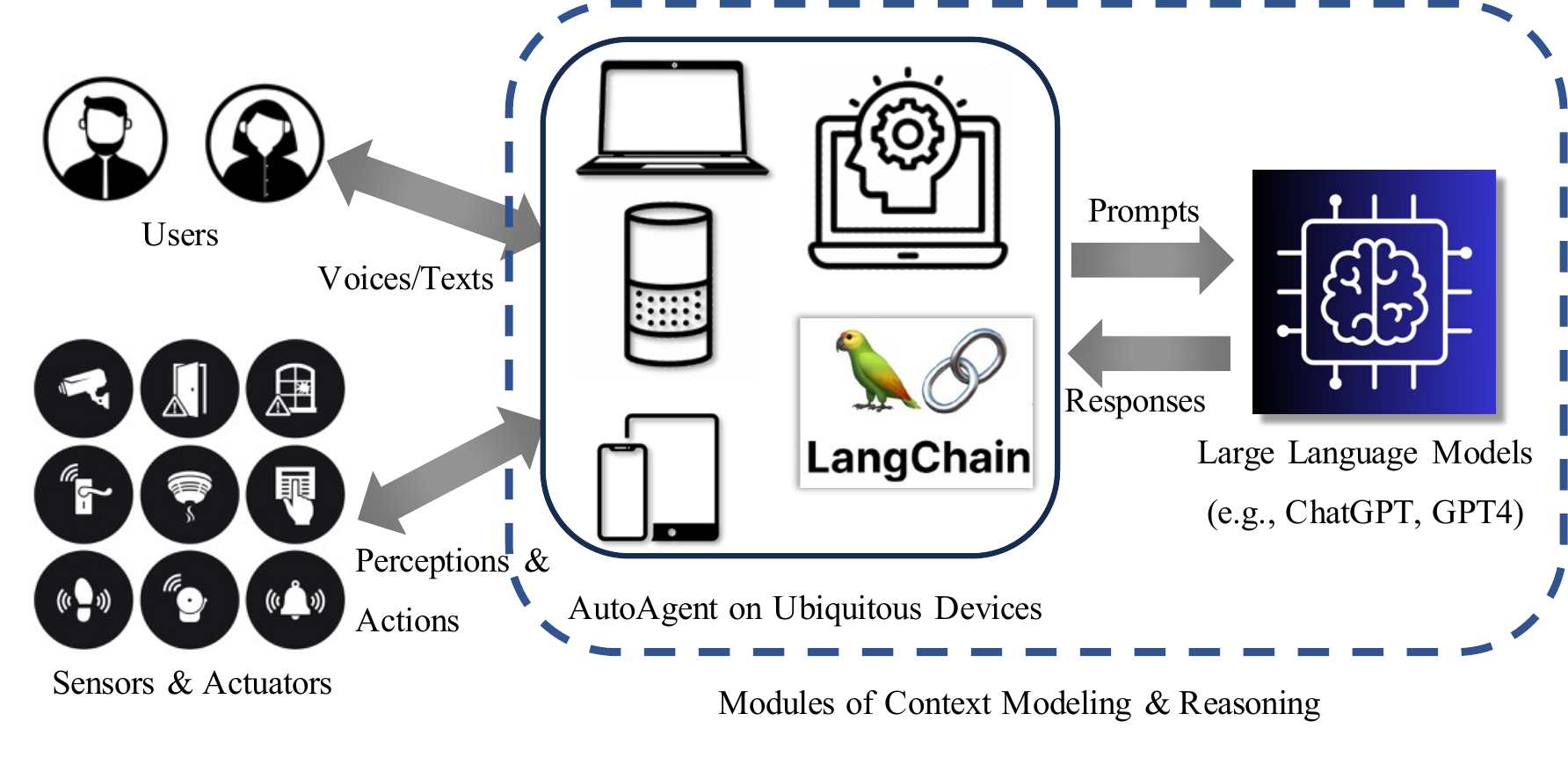}
    \caption{The Architectural Design of LLM-driven Context-aware Computing (\TheName{}) with Users, LLMs, Autonomous Agent (AutoAgent) on Ubiquitous Devices, and Sensors \& Actuators}
    \label{fig:llm-driven}
\end{figure*}

This work organizes and introduces the existing studies laying on the intersection of context-aware computing and LLMs, and further summarizes a new computing paradigm, namely \underline{L}LM-driven \underline{C}ontext-\underline{a}ware \underline{C}omputing (\TheName), which leverages texts, prompts, AutoAgents and LLMs for context modeling and reasoning. As shown in Figure~\ref{fig:llm-driven}, the architectural design of \TheName{} consists of four primary components: Users, Sensors \& Actuators, AutoAgents, and LLMs. Note that, inspired by~\cite{franklin1997autonomous},  we assume AutoAgents are programs running on the Ubiquitous Devices, which are capable of interacting with users, accessing to LLMs and the Web, and communicating with Sensors \& Actuators\footnote{In the rest of manuscript, we may use the term ``AutoAgent'', ``Ubiquitous Device'' and ``AutoAgent on the Ubiquitous Device'' alternatively, as they basically point to the same thing, i.e., the program of autonomous agents running on ubiquitous devices. When we emphasising the function as a hardware equipment, we may prefer the term ``Ubiquitous Device''. Otherwise, when considering it as a program to process data, we may use the term ``AutoAgent''.}.
%
%The latter two components serve as the modules for context modeling and reasoning.
%
The AutoAgent on ubiquitous devices plays a centric role in context modeling \& reasoning, as it converts sensor data into text to represent the context and translates texts into commands to control actuators. Users can interact with the ubiquitous device by sending requests, which are forwarded along with the context to the LLM, such as GPT-4, for context reasoning. The LLM responds with structured text containing actions to fulfill the user's request. Finally, the AutoAgent on ubiquitous devices controls actuators to follow the LLM's response, enabling it to execute the actions and fulfill the user's request. 

This work contributes primarily through a systematic review of existing literature, thereby bridging the gap between ubiquitous computing and AI communities. The review embarks with the introduction of Large Language Models (LLMs) and Generative Pre-trained Transformer (GPT), further providing insights into various LLM-related techniques. We subsequently present the \TheName{} architecture that embodies an integrative solution using LangChain and AutoAgents. The \TheName{} system distinguishes itself through the critical role of AutoAgent in modeling context, performing context reasoning, and commanding actuators based on LLM response, elucidated across two showcases in medication and trip planning. Furthermore, we explore key performance factors with emphasis on token limits, prompting techniques for action planning, trustworthiness and interpretability of LLMs, domain-specific tuning, task decomposition, and the use of physical embodiment for AutoAgents. We conclude with a projection of future research directions, emphasizing the need for integrative models, advanced transformer architectures, and efficient task decomposition to enhance the performance of LLM-powered autonomous agents. 

%Note that, comprehensive tutorials on context-aware systems design could be in~\cite{dey2001conceptual,dey2001understanding}. The history as well as the progress have been reviewed in the Chapter of \emph{Context-aware Computing} of~\cite{krumm2018ubiquitous} by Anind K Dey. On the other hand, \emph{Bubeck et al.}~\cite{bubeck2023sparks} provides an technical review on the capacity of LLMs for AGI. The online review~\cite{lilian2023LLM} present the most recent technologies and practices on LLM-powered AutoAgents. However, as an emerging research direction, few efforts have been done on the intersection between Context-aware Computing and LLM-powered AutoAgents. Compared to existing reviews and tutorials, our work should be the first tutorial, aiming to fill the gap between context-aware computing and large language model communities, with examples, reviews, and discussions.

Note that comprehensive tutorials on context-aware systems design can be found in the earlier works from \emph{Dey et al.}~\cite{dey2001conceptual,dey2001understanding}. The history and progress of this field have been reviewed in the chapter on Context-aware Computing in~\cite{krumm2018ubiquitous}, also by Anind K. Dey in 2018. On the other hand, Bubeck et al. ~\cite{bubeck2023sparks} provide a technical review of the potential of LLMs for AGI. The online review by Lilian Wang~\cite{lilian2023LLM} presents the most recent technologies and practices in LLM-powered AutoAgents. However, as an emerging research direction, there have been few efforts at the intersection of Context-aware Computing and LLM-powered AutoAgents. In comparison to existing reviews and tutorials, our work aims to be the first tutorial that fills the gap between context-aware computing and large language model communities, offering examples, reviews, and discussions.

The manuscript is organized as follows. In Section 2, we review the background and preliminary works in the field. Section 3 presents the architectural design of \TheName{} and its LangChain-based implementation. Section 4 demonstrates two showcases as the proof-of-the-concept. Section 5 elaborates the key factors that may affect the performance of \TheName{} and discuss the future research directions. Finally, we conclude this work in Section 6.

\section{Backgrounds and Preliminaries}
In this section, we brief the preliminary studies related to large language models (LLMs) and generative pre-trained transformers (GPT), especially we focus on introducing novel LLM-driven techniques, including question answering (QA), prompt-based tuning, and instruction following, that could be adopted for context modeling and reasoning.

\subsection{Foundation Models, Large Language Models (LLMs) and Generative Pre-trained Transformers (GPT)}
\begin{figure}
    \centering
    \includegraphics[width=0.65\textwidth]{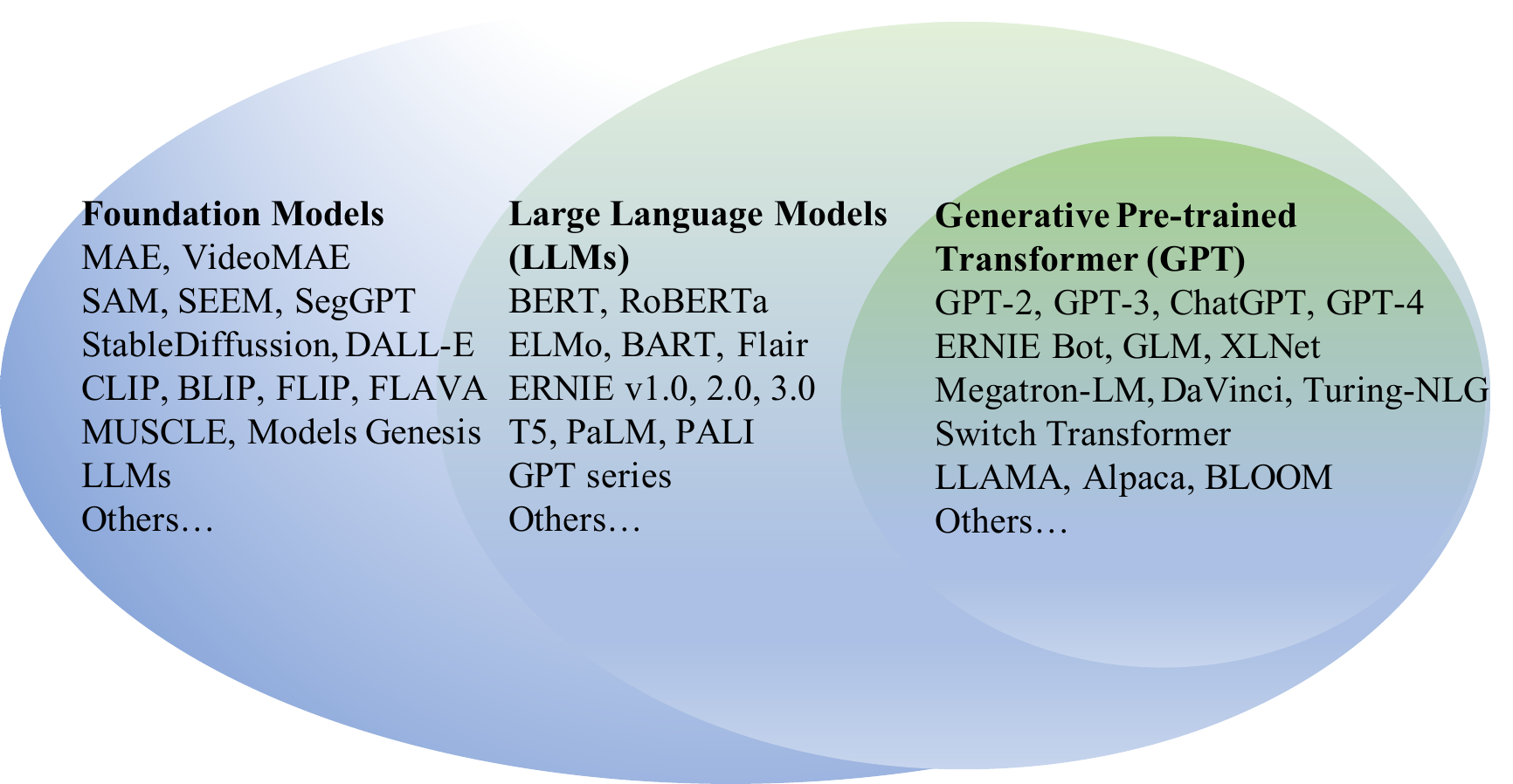}
    \caption{Concepts of Foundation Models, Large Language Models (LLMs), and Generative Pre-trained Transformer (GPT)}
    \label{fig:concepts}
\end{figure}

In this section, we introduce the concepts of \emph{Foundation Models}, \emph{Large Language Models} (LLMs), and \emph{Generative Pre-trained Transformers (GPT)}. Figure~\ref{fig:concepts} illustrates the nested relationships of these three concepts.

\subsubsection{Foundation  Models}
Foundation models act as the base architecture in deep learning and  have been considered as a ``paradigm for building AI systems'' for either perception or generation purposes~\cite{murphy2021foundation}. They are often trained on large amounts of unlabeled/labeled data in a self-supervision manner, and are designed to extract and learn complex patterns from data. Once trained, these models can then be fine-tuned or adapted for specific downstream tasks for application purposes.

Primarily, foundation models could be categorized by their supporting data modalities, such as natural language processing (NLP)~\cite{zhou2023comprehensive}, computer visions~\cite{liu2023survey,khan2022transformers}, acoustic \& speech~\cite{hadian2023review}, multi-modalities~\cite{du2022survey,wang2023large}, and AI for Sciences~\cite{anstine2023generative}. In the meanwhile, from the perspectives of tasks, foundation models could be categorized as perception models and generation models~\cite{zhang2023complete,anstine2023generative}. summary, foundation models provide the necessary structure and learning capabilities to understand and generate useful outputs from complex and diverse data types across different modalities~\cite{bommasani2021opportunities,murphy2021foundation}.

\subsubsection{Large Language Models}~LLMs represent a cutting-edge approach to natural language processing, enabling machines to better understand and generate human-like texts. Give the superb language capacity, LLMs have been used to perform various tasks, such as text summarization, machine translation, sentiment analysis, and question-answering, thereby revolutionizing the way human and computer interact. In terms of LLMs design, three primary architectures have been invented: encoder-only, encoder-decoder, and decoder-only. Encoder-only architectures, such as BERT~\cite{bert} or ERNIE~\cite{ernie}, focus on understanding the input context and creating meaningful representations, which can be used for tasks like sentiment analysis and document classification. Encoder-decoder architectures, exemplified by models like T5~\cite{raffel2020exploring}, BART~\cite{lewis2019bart} and GLM~\cite{du2021glm}, consist of two components: the encoder processes the input text, while the decoder generates the output. These models are particularly suited for machine translation, summarization, and conversational applications. Decoder-only architectures, like ChatGPT and GPT-4, primarily generate text given some input context, performing tasks such as text completion, story generation, and creative writing~\cite{yang2023harnessing,bubeck2023sparks,zhou2023comprehensive}. Overall, LLMs provide users a powerful tool for navigating and generating human-like text in various contexts.

\subsubsection{Generative Pre-trained Transformer}~Moreover, Generative Pre-trained Transformer (GPT)~\cite{radford2018improving} refers to a series of state-of-the-art LLMs, which usually trained by four stages -- Pre-training, Post Pre-training (PPT), Supervised Fine-Tuning (SFT), and Reinforcement Learning with Human Feedback (RLHF). Among these four stages, two of them are mandatory -- pre-training and  SFT~\cite{chen2020big}. During the (autoregressive) pre-training stage, GPT learns to predict the next word in a sentence given the context of previous words, i.e., completion of sentences or paragraphs, which helps the model acquire a strong grasp of syntax, semantics, and general knowledge. After pre-training, SFT is employed to adapt the model to specific tasks, such as question-answering or sentiment analysis, by leveraging labeled data. In general, pre-training relies on massive collection of texts, while SFT often requires well-annotated question-answering (QA) pairs as labeled data for training.

In addition to above two, Post Pre-training (PPT) and Reinforcement Learning from Human Feedback (RLHF) are two optional stages for GPT training. To customize domain-specific models, it sometimes requires to add one additional post pre-training (PPT)~\cite{pan2020multilingual} procedure between pre-training and SFT phases. Furthermore, Reinforcement Learning from Human Feedback (RLHF)~\cite{christiano2017deep} can further optimize GPT's performance posterior to the SFT, allowing it to learn from human-generated comparisons and adjust its output accordingly. Compared to the two mandatory stages, PPT is also based on autoregressive training tasks but leverages domain-specific collection of texts, so as to adapt the domain shift from general domains (for pre-training) to the application domains, such as healthcare and law~\cite{jo2023understanding,xu2023leveraging,haupt2023ai,bommarito2022gpt}.

\begin{figure*}
    \centering
    \includegraphics[width=\textwidth]{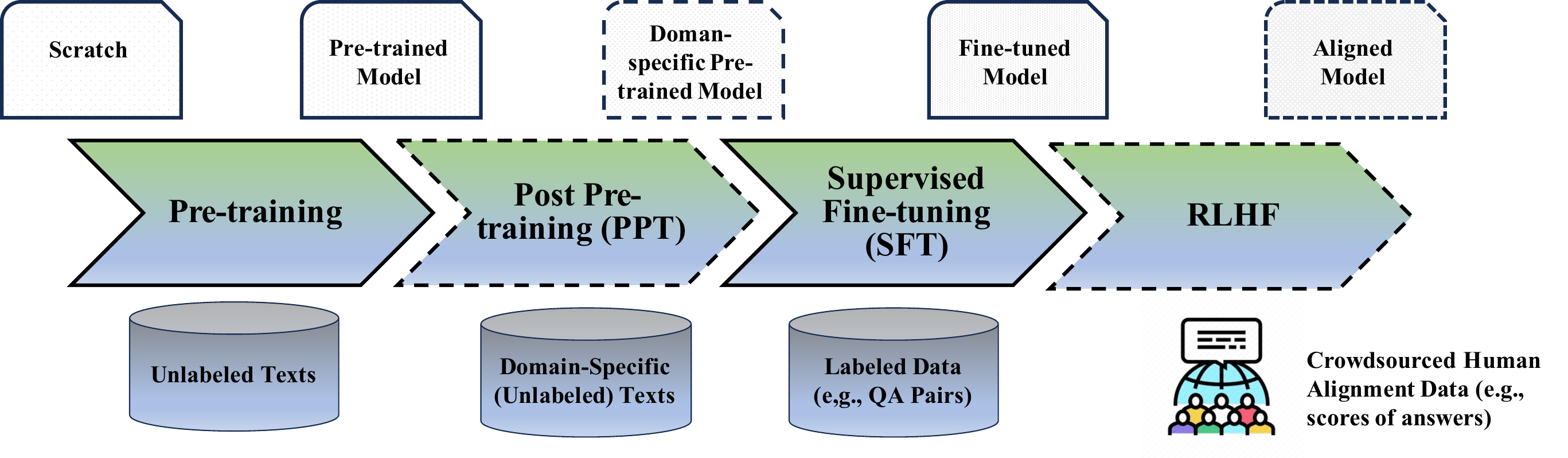}
    \caption{The Training Phases and Outcomes of GPT Training: Pre-training, Post Pre-training (PPT), Supervised Fine-tuning (SFT), and Reinforcement Learning with Human Feedback (RLHF)}
    \label{fig:train-phases}
\end{figure*}

\subsection{Prompts, Prompting, and Prompt-based Tuning}
GPT has showcased remarkable capacities in knowledge retention, question-answering (QA), and instruction following. It can effectively answer questions based on information acquired during pre-training, demonstrating a certain understanding of language and contexts. More specifically, users can input their instructions to GPT in the context of a question, then the model follows the instruction by responding the question. To better respond the questions or accurately follow the instructions, LLMs have been developed to incorporate advanced prompt-based QA techniques as follows.

\subsubsection{Prompting} Furthermore, GPT can be extended with the use of prompts, which are carefully designed input texts (usually with templates) that guide the LLM towards producing the desired output. Specifically, a prompt~\cite{brown2020language} is an input text or phrase that is given to the LLM to initiate a response. The LLM then generates a relevant and coherent continuation of the text based on the prompt. This allows users to makes LLMs suitable for a wider range of applications with natural language.

\begin{figure*}%[!h]
    \centering
    \includegraphics[width=\textwidth]{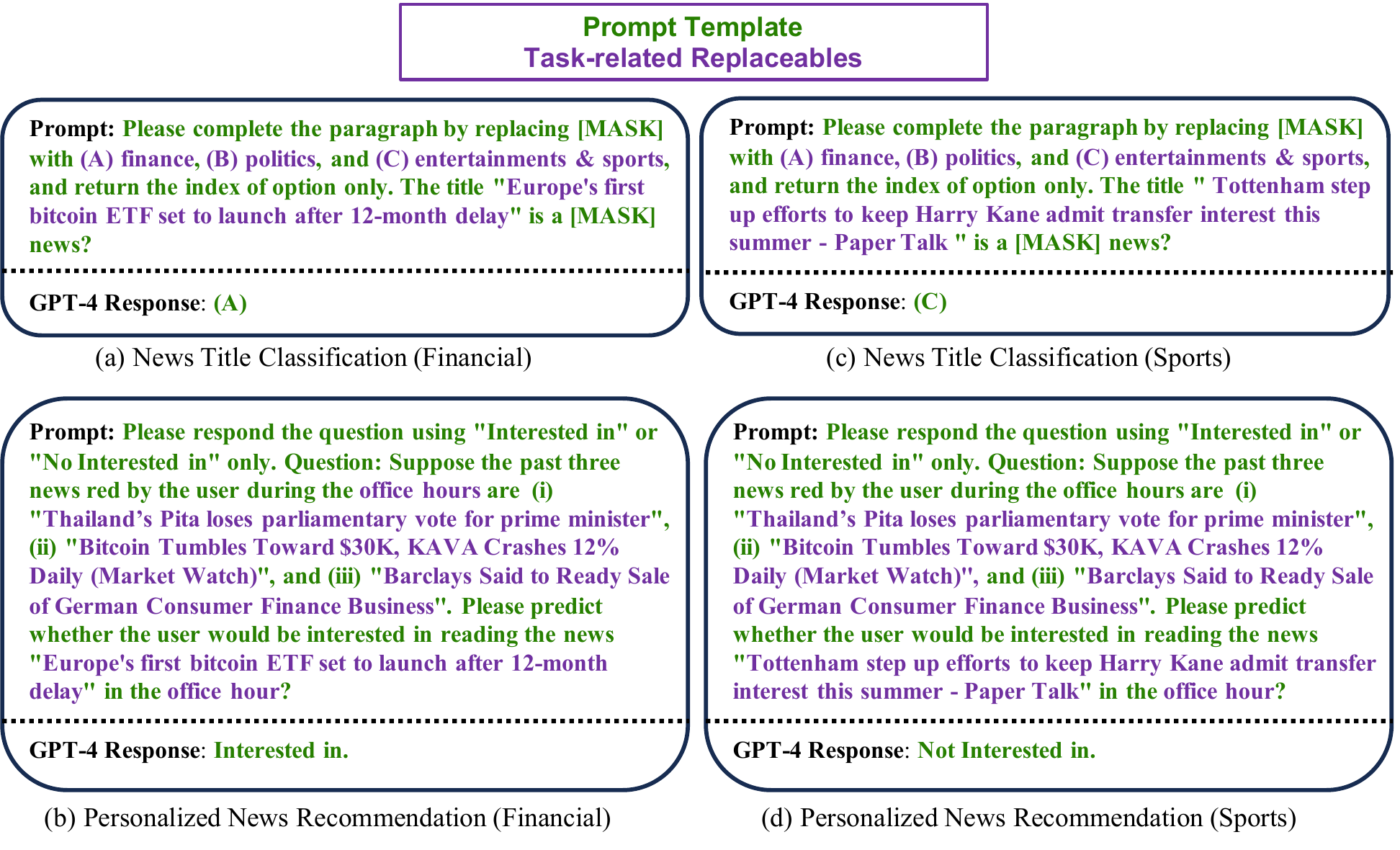}
    \caption{The Prompt and Response for News Title Classification and Personalized Recommendation with GPT-4}
    \label{fig:news-class}
\end{figure*}

\begin{example}[Prompting for News Classification]
an AutoAgent running on the tablet receives a notification of \emph{news} from a mobile feed app, during the office hour of the user. The AutoAgent has to first categorize the type of news, such as finance, politics, entertainments \& sports and so on, and then decides whether the user would be interested in the news. Suppose the notification is as follows.
\begin{itemize}
    \item[] \textbf{News Title:} {``Europe's first bitcoin ETF set to launch after 12-month delay''}.
\end{itemize}
To classify the news, as shown in Fig.~\ref{fig:news-class}~(a), the AutoAgent sends the prompt to GPT-4 and received the response as the results of news title classification.
\end{example}

In above example, the AutoAgent could simply enable GPT-4 as a classifiers of news categories, without any modifications to LLMs, and leverages the response from GPT-4 as the result of classification for decision-making.

\subsubsection{Few-shot Prompting and In-Context Learning}
On top of prompting techniques, few-shot prompting~\cite{zhao2021calibrate} refers to the ability of LLMs to understand and generate accurate responses, just by including a few examples or demonstrations (e.g., question-answer pairs) in the prompt. This is in contrast to traditional deep learning models that often require large amounts of labeled data for training. Few-shot Prompting doesn't require further fine-tuning on LLMs. Let's continue the example of news categorization and recommendation. 

\begin{example}[Few-shot Prompting for Personalized News Recommendation]
With the news notification received, the AutoAgent needs to decides whether it should interrupt the user for the news in a personalized manner~\cite{rosenthal2011using}. In this case, the AutoAgent first retrieves the past records of news reading in the same context. For example, the last three news red by the users, in the office hours, are  (i) ``Thailand’s Pita loses parliamentary vote for prime minister'', (ii) ``Bitcoin Tumbles Toward \$30K, KAVA Crashes 12\% Daily (Market Watch)'', and (iii)``Barclays Said to Ready Sale of German Consumer Finance Business''. Later, as shown in Fig.~\ref{fig:news-class}~(b), the AutoAgent leverages few-shot prompting and uses the responses from GPT-4 to make the decision of recommendation.
\end{example}

%\begin{itemize}
%        \item \textbf{Prompt}: \emph{Please respond the question using ``Interested in'' or ``No Interested in'' only. Question: Suppose the past three news red by the user during the office hours are  (i) ``Thailand’s Pita loses parliamentary vote for prime minister'', (ii) ``Bitcoin Tumbles Toward \$30K, KAVA Crashes 12\% Daily (Market Watch)'', and (iii)``Barclays Said to Ready Sale of German Consumer Finance Business''. Please predict whether the user would be interested in reading the news ``Europe's first bitcoin ETF set to launch after 12-month delay'' in the office hour?}%\vspace{-3mm}
%        \item \textbf{GPT-4 Response}: \emph{Interested In.}
%\end{itemize}

Apparently, based on the user's previous reading habits during office hours, it seems that the user has an special interest in news related to politics, and finance (including cryptocurrency). The news ``Europe's first bitcoin ETF set to launch after 12-month delay'' falls under the category of finance (and cryptocurrency), which the user has shown interest. Therefore, it is likely that the user would be interested in reading this news. In Fig.~\ref{fig:news-class}~(c) and (d), we also provide another example on the sport news that the user might not be interested in reading during the office hour. In above cases, the AutoAgent can use GPT-4 as a learner of personalized news recommendation.  This paradigm also refers to In-Context Learning (ICL)~\cite{dong2022survey}, where the LLM learns from the demonstrations embedded in the prompt for classification and recommendation, without tuning the model.

%More importantly, the use of prompts enables a novel few-shot learning paradigm, namely prompt-based tuning~\cite{lester2021power}, which makes GPT learn to perform a new downstream task from \emph{few examples} but without updating the parameters of the model. 

\begin{algorithm*}%[t!]
  \caption{Pseudo Code of Prompting in the General Form via Question Answering with GPT-4.}
  %\small
  \label{alg:prompt}
 %  \scriptsize
\KwSty{import} \ProgSty{os}\\
\KwSty{import} \ProgSty{openai}\\
\;
\CommentSty{\# Template for prompt-based tuning with placeholders \KwSty{\{0\}}, \KwSty{\{1\}}, \KwSty{\{2\}}}\\
\KwSty{prompt\_template}  =  ``\CommentSty{Please answer the question by considering descriptions }''\;
\KwSty{prompt\_template} += ``\CommentSty{and examples below. \textbackslash n\textbackslash n}'' \;
\KwSty{prompt\_template} += ``\CommentSty{Descriptions: \KwSty{\{0\}}. \textbackslash n }'' \;
\KwSty{prompt\_template} += ``\CommentSty{Examples: \KwSty{\{1\}}. \textbackslash n \textbackslash n }'' \;
\KwSty{prompt\_template} += ``\CommentSty{Question: \KwSty{\{2\}}. \textbackslash n }'' \;
\KwSty{prompt\_template} += ``\CommentSty{Answer: }'' \;
\;
\CommentSty{\# Question answering with texts of} \KwSty{description}, \KwSty{example}, \KwSty{question} \CommentSty{as inputs}\;
\ProgSty{def prompt\_based\_QA(\KwSty{description}, \KwSty{example}, \KwSty{question})}:\;
\hspace{7mm}    \KwSty{prompt} = \KwSty{prompt\_template}.\FuncCall{format}{\;
\hspace{12mm}   \KwSty{description}, \hspace{10mm} \CommentSty{\# Replacing placeholder \KwSty{\{0\}} in the prompt}\; 
\hspace{12mm}   \KwSty{example},     \hspace{18.5mm} \CommentSty{\# Replacing placeholder \KwSty{\{1\}} in the prompt}\; 
\hspace{12mm}   \KwSty{question}} \hspace{18.5mm} \CommentSty{\# Replacing placeholder \KwSty{\{2\}} in the prompt}\\
\hspace{7mm} \CommentSty{\# Function call through OpenAI API to complete the QA}\\
\hspace{7mm}\KwSty{response} = \FuncCall{openai.ChatCompletion.create}{ \;
\hspace{12mm}     \var{model="gpt-4"}, \;
\hspace{12mm}      \var{messages=[{"role": "user", "content": \KwSty{prompt}}]}, \var{other\_hyperparameters ...}
}\\
\hspace{7mm}\Return {\KwSty{response}}
\end{algorithm*}

\subsubsection{Prompting in the General Form}
In above \textbf{Examples 1} and \textbf{2}, we provide two cases based on \emph{zero-shot prompting} and \emph{few-shot prompting}, where different templates of prompts have been used to make LLMs perform as a classifier and recommender systems without further fine-tuning the models. To avoid lose of generality, Algorithm~\ref{alg:prompt} presents the python-style pseudo code of prompt-based tuning for question answering, where the description and examples of tasks are given as the prior of the question. Specifically, the program in Algorithm~\ref{alg:prompt} templates the given texts into a prompt, invokes ChatGPT's API with the prompt to obtain the response from GPT-4~\cite{openai2023gpt4}, and returns the response. On the other side, GPT responds the question by paragraph completion following the prompt template. Later, we will provide showcases using the template in Algorithm~\ref{alg:prompt} for prompting in general-purposes.

\subsubsection{Prompt-based Tuning} 

Prompt-based tuning~\cite{liu2023pre,peft} refers to a set of simple yet effective mechanisms for tuning the ``frozen'' language models to perform specific downstream tasks through optimizing the ``prompts''~\cite{reynolds2021prompt,jiang2020can,shin2020autoprompt,li2021prefix,lester2021power,liu2021gpt,liu2022p}. All these methods aim at fine-tuning LLMs subject to the downstream tasks while keeping the parameters of models frozen. They all need the collection of labeled data based on well-annotated question-answer pairs. To tune the model in a supervised manner, two major types of prompt-based tuning techniques--``discrete pompts'' and ``continuous prompts''--have been introduced~\cite{liu2023pre}.

To optimize ``discrete prompts'', search-based and generator-based methods have been studied~\cite{reynolds2021prompt,jiang2020can,shin2020autoprompt,pryzant2023automatic} for enhancing the input sequences sent to the (frozen) LLM through searching or generating texts. On the other hand,  continuous prompting~\cite{liu2023pre,peft} techniques intend to tune the model by introducing an extremely small number of additional learnable parameters.
Specifically, prefix-tuning~\cite{li2021prefix} propose a small \emph{continuous task-specific vector}, namely prefix, as virtual tokens prior to the input sequences of LLMs, while prompt tuning~\cite{lester2021power} and P-tuning~\cite{liu2021gpt} leverage prompt encoders to enhance tokens in the input sequence for text generations. P-tuning v2~\cite{liu2022p} inserts and optimizes layer-wise prompts for every layer of the LLM, which delivers performance as good as supervised fine-tuning on all parameters in the model. PEFT~\cite{peft} surveys this area and provides open-source implementations of existing parameter-efficient fine-tuning algorithms in Huggingface.

\begin{table}[h] 
\caption{Comparisons among different ``learning from examples'' strategies of LLMs. (Seq2Seq: Sequence to Sequence Prediction, PPT: Post Pre-training; SFT: Supervised fine-tuning; RLHF: Reinforcement Learning from Human Feedback; PPO: Proximal Policy Optimization; ICL: In-Context Learning; QA: Question Answer; and ContP: Continuous Prompts)}\label{tab:comparision-learn}
\begin{tabular}{r|c|c|c} \hline 
 & \textbf{Training Objectives} & \textbf{Training Data} & \textbf{Model Weight Updates} \\ \hline 
Pre-training~\cite{radford2018improving} & Autoregressive & Texts (general)& Gradients (LLM)\\ %\hline 
PPT~\cite{pan2020multilingual} & Autoregressive & Texts (domain)& Gradients (LLM)\\  %\hline 
SFT~\cite{chen2020big} & Seq2Seq & QA Pairs& Gradients (LLM)\\  \hline 
\multirow{2}{*}{RLHF~\cite{ziegler2019fine,lambert2022illustrating}} & PPO~\cite{schulman2017proximal} & QA Scores & Gradients (Reward Model)\\ 
&  Seq2Seq~(SFT) & Generated QA & Gradients (LLM) \\\hline 
\multirow{2}{*}{ICL~\cite{dong2022survey}} & \multirow{2}{*}{Unsupervised} & Demonstrations & \multirow{2}{*}{None}\\ %\hline 
&&in prompts&\\ \hline
Template Search~\cite{wallace2019universal,shin2020autoprompt} & Seq2Seq& QA Pairs & None (Token Search) \\ %\hline 
P-tuning~\cite{liu2022p} & Seq2Seq & QA Pairs & Gradients (ContP)\\ %\hline 
Prompt Tuning~\cite{lester2021power} & Seq2Seq & QA Pairs & Gradients (ContP)\\ %\hline 
Prefix Tuning~\cite{li2021prefix} & Seq2Seq & QA Pairs & Gradients (ContP as Prefix Tokens)\\ %\hline 
P-tuning v2~\cite{liu2022p} & Seq2Seq & QA Pairs & Gradients (Layer-wise ContP)\\ %\hline 
Hybrid Tuning~\cite{zhong2021factual,han2022ptr}& Seq2Seq  & QA Pairs & Gradients (ContP) + Token Search\\ %\hline
Prompt Generation~\cite{gao2021making,ben2021pada} & Seq2Seq & QA Pairs& Gradients (Prompt Generator) \\\hline
\end{tabular} 
\end{table}

\subsubsection{Discussions} 
Table~\ref{tab:comparision-learn} provides a concise comparison of the various ``learning from examples'' strategies utilized in Large Language Models (LLMs). These strategies exhibit variations in terms of training objectives, training data, and the methodology of applying model weight updates.

In summary, Pre-training and Post Pre-training (PPT) both leverage an autoregressive training objective with gradient-based model weight updates, differing in their use of general texts and domain-specific texts for training respectively. Supervised fine-tuning (SFT) utilizes Sequence to Sequence (Seq2Seq) prediction with question-answer pairs as training data and updates model weights using gradients or back-propagation. Reinforcement Learning from Human Feedback (RLHF) applies Proximal Policy Optimization (PPO) and Seq2Seq (via SFT) using question-answer scores and pairs. In-Context Learning (ICL) stands unique with an unsupervised approach utilizing demonstrations in prompts for training, without performing model weight updates. Various types of prompt tuning and template search approaches like P-Tuning, Prefix tuning, and Hybrid tuning employ a Seq2Seq prediction framework with QA pairs as training data but differ in the way they update model weights.

%Prompt-based tuning~\cite{lester2021power} is a method used to fine-tune AI language models to generate more accurate and relevant responses to specific prompts. It involves training the model on a dataset containing various prompts and their corresponding desired responses. This helps the model learn patterns and relationships between prompts and responses, improving its ability to generate appropriate answers.
%In summary, prompts are used to guide AI language models in generating relevant responses. Prompt-based tuning helps improve the model's accuracy for specific types of prompts, while few-shot prompting demonstrates the model's ability to learn and respond accurately based on limited examples.

%Furthermore, these prompts could be self-adaptive and learnable subject to the GPT model and the input question~\cite{liu2022p}.

%\subsection{LLM-powered AutoAgents for Embodied AI}

%As early as 1990s, autonomous agents (AutoAgents) – programs that operate independently, make decisions and perform actions based on their environment and goals – have been anticipated to learn and adapt from their experiences in numerous application domains~\cite{franklin1996agent}. However, their practical deployment has been frequently limited~\cite{albrecht2018autonomous}. 

%The surge of large language models (LLMs) and LanghChain has reignited interest in these autonomous agents (AutoAgent)~\cite{lilian2023LLM}. 

%Please follow the online tutorials for details~\cite{langchain2023,lilian2023LLM}.

\subsection{LLM-powered Applications for Ubiquitous Computing and Computer-Human Interactions}
%LLM-powered applications are reshaping ubiquitous computing and computer-human interactions, enabling more natural, personalized, and contextually relevant interactions across a wide range of domains. From mental health support to gaming, journalism, coding assistance, and beyond, LLMs are driving transformative advancements in how we engage with technology in our everyday lives. We summarize popular applications in terms of domain designed in Table~\ref{tab:llm-apps}.

Advancements fueled by LLMs are not tied to the natural language processing domain only. Instead, one can observe a broad landscape of impacts ranging from mental health support~\cite{xu2023leveraging} to gaming~\cite{ashby2023personalized}, journalism~\cite{petridis2023anglekindling}, play scripting~\cite{mirowski2023co}, and coding assistance~\cite{mcnutt2023design} among others. LLMs, characterized by their ability to understand a wider range of human inputs, are significantly enhancing the level of engagement between humans and computing technology.
Table~\ref{tab:llm-apps} provides a comprehensive review on recent studies on LLMs-powered applications for ubiquitous computing and human-computer interaction, published by IMWUT and SIGCHI. 

Specifically, several studies have shown the potential of leveraging sophisticated tools and methodologies for more effective health-related applications. Xu et al. \cite{xu2023leveraging} demonstrated the use of LLMs for predictive mental health diagnostics using online text data. Meanwhile, Rawassizadeh et al. \cite{rawassizadeh2023odsearch} developed ODSearch, a resource-efficient, on-device search engine designed for optimal data retrieval from fitness trackers. Jo et al. \cite{jo2023understanding} explored the application of LLMs in conversational AI for the execution of public health interventions.

\begin{table}[h]
    \centering
    \caption{LLM-powered Applications for Ubiquitous Computing}
    \label{tab:llm-apps}
    \begin{tabular}{p{3cm} p{3.5cm} p{7cm}}
        \toprule
        \textbf{Teams} & \textbf{Highlights} & \textbf{Summaries} \\
        \midrule
        \multicolumn{3}{c}{Health-related Applications}\\ \midrule
        Jo et al.~\cite{jo2023understanding} & Public Health & Conversational AI and LLMs for interventions. \\ 
        Rawassizadeh et al.~\cite{rawassizadeh2023odsearch} & Fitness & Natural language search for fitness tracking. \\
        Xu et al.~\cite{xu2023leveraging} & Mental Health & LLMs for mental health prediction based on online texts. \\
        \midrule\multicolumn{3}{c}{Creative Writings and Related Applications}\\ \midrule
        Petridis et al.~\cite{petridis2023anglekindling} & Journalism & LLMs for journalistic angle ideation. \\
        Dang et al.~\cite{dang2023choice} & Storytelling &  Diegetic and non-diegetic prompts for writing. \\
        Ashby et al.~\cite{ashby2023personalized} & Games & LLMs for game quest and dialogue generation. \\
        Mirowski et al.~\cite{mirowski2023co} & Scripting & LLMs for screenplays and theatre scripts co-writing. \\
        McNutt et al.~\cite{mcnutt2023design} & Code Assistants & LLMs for coding assistant on notebooks. \\
        Wang et al.~\cite{wang2023popblends} & Concepts \& Abstracts & LLMs for abstract thinking and conceptual blending. \\
        Zhou et al.~\cite{zhou2023synthetic} & Misinformation & Evaluating LLMs-generated misinformation. \\ 
        \midrule\multicolumn{3}{c}{User Interaction/User Experience (UI/UX)}\\ \midrule
        Wang et al.~\cite{wang2023enabling} & Mobile UI & Conversational UI for mobile devices with LLMs. \\
        Valencia et al.~\cite{valencia2023less} & AAC Communications & LLMs for impede communication with AAC users. \\
        Hamalainen et al.~\cite{hamalainen2023evaluating} & User studies & LLMs for synthetic HCI research data generation. \\
        Jones et al.~\cite{jones2023embodying} & Arts \& Performance & Artistic performance understanding with LLMs. \\
        \bottomrule
    \end{tabular}
\end{table}

%Xu et al. \cite{xu2023leveraging} have notably delved into the application of large language models for mental health predictions. They showcased the capacity of these models in utilizing online text data as an effective behavioral measure, providing a substantial step forward in mental health diagnostics. Similarly, Rawassizadeh et al. \cite{rawassizadeh2023odsearch} put forth an innovative solution in the form of ODSearch. The tool serves as a resource-conserving, on-device natural language search engine. This advancement is specifically designed to facilitate more efficient data retrieval from fitness trackers, investigating the intersection of data management and physical health. Furthermore, the research efforts of Jo et al. \cite{jo2023understanding} have been directed towards the studies of conversational AI. Backed by LLMs, their work explores the potential of these systems in executing public health interventions.   

%Xu et al.\cite{xu2023leveraging} provided insight into how large language models can be leveraged for mental health prediction, using online text data as a behavioral measure. Rawassizadeh et al.\cite{rawassizadeh2023odsearch} developed ODSearch, a resource-efficient, on-device natural language search engine for efficient data retrieval from fitness trackers. Jo et al.\cite{jo2023understanding} explored deploying conversational AI, powered by LLMs, for public health interventions. 

In terms of creativity writings and communications, various research has highlighted the remarkable application of LLMs. Petridis et al. \cite{petridis2023anglekindling} introduced AngleKindling, a tool leveraging LLMs for unique journalistic angles. Meanwhile, Mirowski et al. \cite{mirowski2023co} used LLMs for a collaborative approach to scriptwriting, blending human creativity with AI, evaluated by professionals. Dang et al. \cite{dang2023choice} probed into how users work with LLMs in writing scenarios, considering different types of prompts. In the gaming industry, Ashby et al. \cite{ashby2023personalized} discussed the integration of LLMs to create personalized quests and dialogues in role-playing games. For code developing, McNutt et al. \cite{mcnutt2023design} unveiled AI-assisted code assistants for notebooks, simplifying code generation and debugging tasks. Wang et al. \cite{wang2023popblends}, employing a more abstract perspective, devised strategies for conceptual blending using LLMs, fostering unique knowledge synthesis pathways. Furthermore, Zhou et al. \cite{zhou2023synthetic} examined the problematic area of AI-generated misinformation, offering possible solutions and emphasizing the ethical responsibilities associated with AI usage.

%Petridis et al.\cite{petridis2023anglekindling} utilized LLMs to assist in journalistic angle ideation through a tool named AngleKindling. Mirowski et al.\cite{mirowski2023co} developed a collaborative approach to writing screenplays and theatre scripts, involving language models evaluated by industry professionals. Valencia et al.\cite{valencia2023less} analyzed how LLMs can enhance or impede communication for Augmentative and Alternative Communication (AAC) users. Ashby et al.\cite{ashby2023personalized} discussed the use of LLMs for personalized quest and dialogue generation in role-playing games, backed by knowledge graphs. Meanwhile, McNutt et al.\cite{mcnutt2023design} utilized AI to design code assistants for notebooks. In conceptual blending, Wang et al.\cite{wang2023popblends} discussed strategies for its execution using LLMs; and Zhou et al.\cite{zhou2023synthetic} conducted research on AI-generated misinformation, studying the misinformation and evaluating potential solutions.

Within the spheres of User Interactions (UI) and User Experiences (UX), several studies have delved into the potential of LLMs. Wang et al. \cite{wang2023enabling}, meanwhile, demonstrated LLMs' potential in aiding conversational interaction within mobile UI. On the other hand, Valencia et al. \cite{valencia2023less} studied how LLMs could enhance or potentially impair communication for Augmentative and Alternative Communication (AAC) users. Additionally, Hamalainen et al. \cite{hamalainen2023evaluating} looked into capacity of LLMs for generating synthetic Human-Computer Interaction (HCI) research data, shedding light on artificial intelligence's increasing significance in HCI research. Finally, Jones et al. \cite{jones2023embodying} took a creative plunge, investigating human-LLM relationships through artistic performance, offering a unique perspective on human-AI interaction.

In summary, LLM-powered applications have been pushing back the frontiers of ubiquitous computing and human-computer interaction research. These applications majorly leverage the exceptional capabilities of LLMs, paving the way for more natural, personalized, and context-relevant interactions across an expansive array of domains.

\section{LLM-Driven Context Modeling and Reasoning}
In this section, we introduce the architectural design and LangChain-based integrative implementation of \TheName{}.

\begin{figure*}
    \centering
    \includegraphics[width=\textwidth]{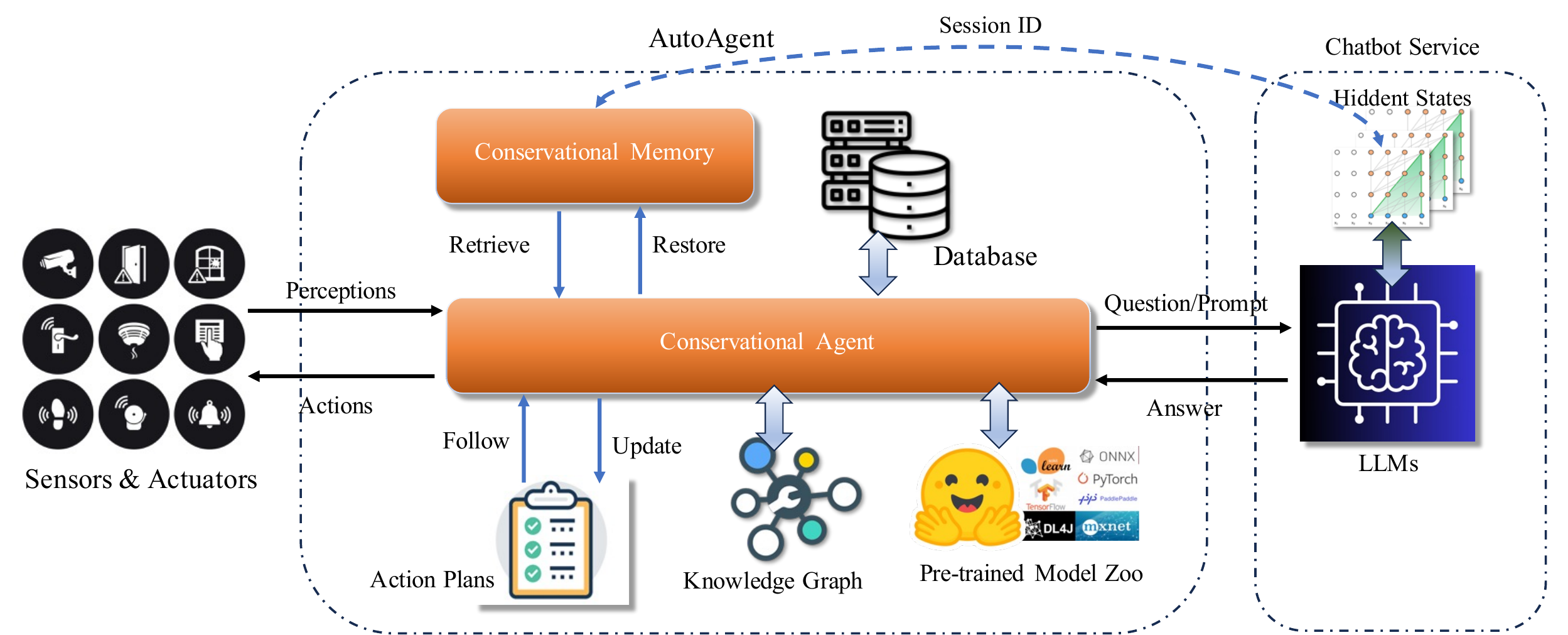}
    \caption{The LangChain-based Integrative Implementation of an AutoAgent with the Conversational Agency, Conversational Memory, Action Plan and External Tool-Uses, such as Databases, Knowledge Graphs, and Pre-trained Model Zoo.}
    \label{fig:ai-agent}
\end{figure*}

\subsection{\TheName{} Architectural Design: An AutoAgent-centric Approach}
In Fig.~\ref{fig:llm-driven}, we present the \TheName{} architecture consists of four primary components: users, sensors \& actuators, an AutoAgent on ubiquitous devices, and LLMs. All these components are connected and integrated through the AutoAgent on ubiquitous devices, while LLMs power the AutoAgent for context modeling and reasoning.

A user communicates with the ubiquitous device via voice or texts, using natural language as a medium to represent instructions and responses. Upon receiving an instruction from a user, the AutoAgent on the ubiquitous device converts the voice input to text and formats the instruction into a prompt, following a predefined template. The AutoAgent then calls the \emph{Question Answering} APIs of the LLM and forwards the prompt as a question to answer.

As introduced in the backgrounds and preliminaries section, a LLM is capable of answering questions with its reasoning, planning, and scheduling abilities~\cite{bubeck2023sparks}. In this way, the LLM is expected to respond the AutoAgent with an answer consisting of actions to implement the user's requests. To facilitate the communication between the LLM and the AutoAgent, the response texts should have been structured in a predefined format. Once the response is received, the AutoAgent decodes it into an action plan. Consequently, AutoAgent follows the action plan and executes these actions, commanding and controlling sensors \& actuators to fulfill the user's request.

Note that, the implementation of a single request may involve multiple rounds of interactions between the LLM and the AutoAgent. This continuous communication ensures that the user's request is accurately understood and executed. The multi-round interactions between AutoAgent on ubiquitous devices and the LLM relies on abilities of multi-round conversations. However, vanilla LLMs are essentially stateless and cannot remember previous conversations. In this case, a \TheName{} system usually the AutoAgent memorizing the state of multi-round conversations.

\subsection{Implementation of an AutoAgent with LangChain}

As early as 1990s, autonomous agents (AutoAgents) – programs that operate independently, make decisions and perform actions based on their environment and goals – have been anticipated to learn and adapt from their experiences in numerous application domains~\cite{franklin1996agent}. However, their practical deployment has been frequently limited~\cite{albrecht2018autonomous}. 

The surge of large language models (LLMs) and LanghChain has reignited interest in these autonomous agents (AutoAgent)~\cite{lilian2023LLM}. In the proposed architecture, we leverage the LangChain to implement the AutoAgent and integrate all other components, including users, sensors\& actuators, LLMs and beyond, for a \TheName{} system. As shown in Fig.~\ref{fig:ai-agent}, a LangChain usually consists of three mandatory components, including a \emph{Conversational Agent}, a \emph{Conversational Memory}, and an \emph{Action Plan}, to meet the essential goals, e.g., multi-round communications, action planning and execution, of an AutoAgent. Further, a LangChain frequently mounts optional tool-uses, such as {databases}, {knowledge graphs}, and the {pre-trained model zoo}, to improve its accuracy of context modeling \& reasoning while extending its capacity of actions.

\begin{algorithm*}%[t!]
  \caption{An Example of the Conversational Agent via Prompt-based Question Answering}
  %\small
  \label{alg:ai-agent}
 %  \scriptsize
\KwSty{import} \ProgSty{os}\\
\KwSty{import} \ProgSty{openai}\\
...\;
\CommentSty{\# One-shot prompting for the user's request classification}\\
\KwSty{desc\_req\_type}~~=``\CommentSty{Read the user's request in the Question and categorize it}'' \;
\KwSty{desc\_req\_type}+=``\CommentSty{using one of following types.\textbackslash n}''\\
\KwSty{desc\_req\_type}+=``\CommentSty{(A) take medicine, (B) appliance  control, (C) food \& }''\\
\KwSty{desc\_req\_type}+=``\CommentSty{beverage...\textbackslash n Please answer the index of option only.}''\\
\KwSty{examp\_req\_type}~~=``\CommentSty{\textbackslash n**********\textbackslash n}'' \\
\KwSty{examp\_req\_type}~~=``\CommentSty{\textbackslash n Question: turn on the heater when the temperature is below}'' \\
\KwSty{examp\_req\_type}~~=``\CommentSty{freezing\textbackslash n Answer: (C) \textbackslash n}'' \\
\KwSty{examp\_req\_type}+=``\CommentSty{\textbackslash n**********\textbackslash n}'' \\
...\;
\ProgSty{while(\KwSty{agent\_is\_running})}:\;
\hspace{7mm} \KwSty{request}=\FuncCall{wait\_and\_receive}{}\;
\CommentSty{\# Classify the type of user's request, and match the prompt template}\;
\hspace{7mm} \KwSty{req\_type}=\FuncCall{prompt\_based\_QA}{\KwSty{desc\_req\_type},~\KwSty{examp\_req\_type},~\KwSty{request}}\;
\hspace{7mm} \KwSty{desc\_req},~\KwSty{examp\_req}=\FuncCall{match\_prompt\_template}{\KwSty{req\_type}}\;
\CommentSty{\# Read all sensors' data and enhance the context-aware description}\;
\hspace{7mm} \KwSty{sensor\_data}=\FuncCall{read\_sensors}{\FuncCall{get\_all\_sensors}{}}\;
\hspace{7mm} \KwSty{ctx\_desc\_req}=\FuncCall{context\_aware\_description}{\KwSty{req\_type},~\KwSty{sensor\_data}}\;
\CommentSty{\# Call the LLM to make an action plan for context reasoning}\;
\hspace{7mm} \KwSty{response} = \FuncCall{prompt\_based\_QA}{\KwSty{ctx\_desc\_req},~\KwSty{examp\_req},~\KwSty{request}}\;
\hspace{7mm} \KwSty{action\_plan} = \FuncCall{make\_action\_plan}{\KwSty{response}}\;
\CommentSty{\# Follow the action plan for enabling context-awareness}\;
\hspace{7mm} \FuncCall{follow\_actions}{\KwSty{action\_plan},~\FuncCall{get\_all\_sensors}{},~\FuncCall{get\_all\_actuators}{}}\;
...\;
\end{algorithm*}

\subsubsection{Conversational Agent} Upon the user's request, the \emph{Conversational Agent} in a LangChain first initiates conversations with the LLM, then sends questions/prompts to the LLM, which models the context, performs context reasoning, and responds accordingly. Based on the response, the agent makes/updates the action plan as the result of context reasoning, and follows the plan to fulfil the user's request. When following the action plan, the agent may need to communicate with sensors \& actuators for the purposes of perceptions and actions. On the other hand,  the \emph{Conversational Memory} restores previous conversations between the \emph{Conversational Agent} and the LLM and compresses them as the ``states'' of multi-round conversations, while the \emph{Conversational Agent} leverages the state information to improve the accuracy and relevance of multi-round conversations, action planning, and context modeling \& reasoning.

Algorithm~\ref{alg:ai-agent} illustrates the pseudo code for a simple example of the \emph{Conversational Agent} that first receives the request from the user, then categorizes the request using a one-shot prompting-based request classifier. Specifically, given the description and examples for the user's request classification task (lines 4--12 of Algorithm~\ref{alg:ai-agent}), the request classifier works similarly as the news title classifier in \textbf{Examples 1} and \textbf{2}, while responding the agent with the type of request. With the request type, the agent matches the suitable template (including the texts of task description and examples) from the repository of prompt templates~\cite{langchain2023} (lines 16--18). Later, the agent reads all sensors, converts the readings as the texts of contexts, then enhances the task description with contexts (lines 19--21). Finally, the agent packages the context-enhanced task description, examples and the user's request into a prompt, forwards the prompt to the LLM (lines 22--24). The LLM responds the agent with the texts describing an action plan as the results of context reasoning, while the agent follows the plan by incorporating sensors \& actuators for perceptions and actions(lines 25--26). In this way, the \emph{Conversational Agent} plays a central role integrating the necessities of a \TheName{} system in a LangChain.

\subsubsection{Conversational Memory} A way to implement the memory in a LangChain is incorporating previous conversations as part of inputs to the LLM. However, the length of texts often exceeds the token limit after multiple rounds of conversations. Given a LLM with a fixed-length input, one can use the pre-computed hidden-states (i.e., key and values in the self-attention blocks of the transformer) of the last conversation as the memory~\cite{dai2019transformer}. For example, one can retrieve the hidden state of a Huggingface transformer~\cite{model-outputs} by accessing the field \texttt{past\_key\_values}. Actually, the hidden states are stored at the server of LLMs (Chatbot service), which creates, stores and manages a session for every conversation of multiple rounds in a non-volatile way. Hereby, the \emph{Conversational Memory} in a LangChain could restore the identities of all sessions to manage the states of multi-round conversations, while the \emph{Conversational Agent} uses the session ID to track and continue the conversation via API calls.

In fact, the WebApp of ChatGPT and GPT-4 uses independent sessions to manage every multi-round conversations between a user and the LLM, where the user can pickup any session to continue the conversation in the WebApp. However, the current OpenAI's chat completion API, demonstrated in Algorithms~\ref{alg:prompt} and~\ref{alg:ai-agent}, doesn't directly support continuation of the conversation though associating with session IDs~\cite{session-id}. To keep the session for continuation of the multi-round conversations with ChatGPT, one can store the texts of past conversations in a database attached to the LangChain, while appending the texts of past conversations~\cite{keep-session} in each round of new conversation. Otherwise, one can deploy their own LLMs~\cite{touvron2023llama,du2021glm} while leveraging hidden states to manage the sessions.

\subsubsection{Action Plan} To fulfill the user's request, especially for some tedious tasks under complex contexts, an AutoAgent makes the plan as a trajectory of actions~\cite{wei2022chain,yao2023tree,yao2022react,liu2023chain,laskin2022context,shinn2023reflexion}. Key challenges here include the lack of domain-specific, token limits of input/output sequences of LLMs for plan-making, and abilities of optimal planning \& scheduling. Specifically, Chain of Thoughts~\cite{wei2022chain} and Tree of Thoughts~\cite{yao2023tree} guide the LLM to think step-by-step either ``linearly'' or in a tree-structure, so as to facilitate the task decomposition. In addition, techniques like Chain of Hindsight (CoH)~\cite{liu2023chain} and Algorithm Distillation (AD)~\cite{laskin2022context} help in learning from past outcomes to optimize the trajectory of actions. More recently, LLM+P~\cite{liu2023llm+} studies to incorporate external domain-specific planners, while Reflexion~\cite{shinn2023reflexion} proposes verbal reinforcement learning, both for advanced plan-making.

\begin{figure*}
    \centering
    \includegraphics[width=\textwidth]{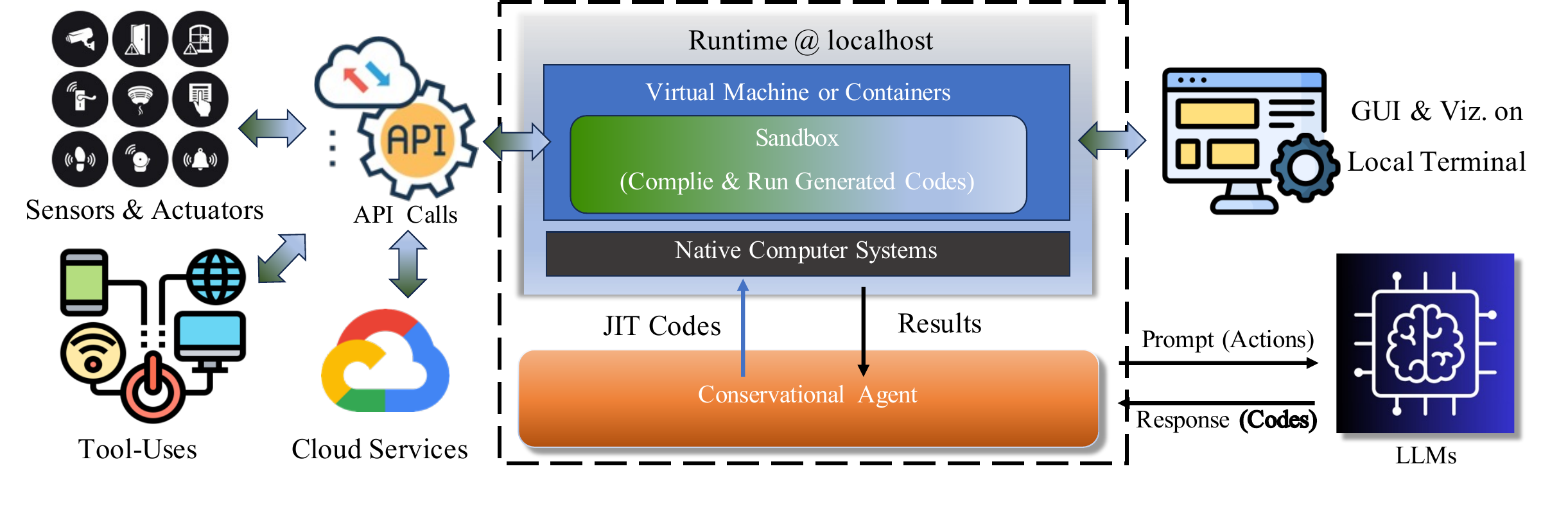}
    \caption{``Just-in-Time'' (JIT) Codes Generation and Execution for Action Planning and Implementation}
    \label{fig:code-runtime}
\end{figure*}

Give the actions represented as texts, AutoAgent of \TheName{} could generate programming codes with the texts, then compile and run the codes to implement the actions. As shown in Fig.~\ref{fig:code-runtime}, the Conversational Agent include the action plan in the prompts sending to the LLM. Later, the LLM responds the agent with ``Just-In-Time'' (JIT) codes~\cite{aycock2003brief}, which implement the actions using program codes. The Agent then uploads JIT codes onto the runtime environment (e.g., a virtual machine or container) at a local ubiquitous device for code compilation and execution. Note that the JIT codes are usually complied and executed in an isolated environment, such as running in a sandbox, for trustworthiness issues. The JIT codes could access external utilities, including sensors \& actuators, cloud services, and other tool-uses through API calls. In addition, for the ubiquitous device with user interaction functionalities, the JIT codes could launch user interfaces and visualization tools for advanced user interactions. RunGPT is such an extension of ChatGPT\footnote{RunGPT: ChatGPT executor - \url{https://rungpt.online/}} for running the GPT-generated codes at the local devices.

\subsubsection{Tool-Uses} 
\begin{figure*}
    \centering
    \includegraphics[width=0.9\textwidth]{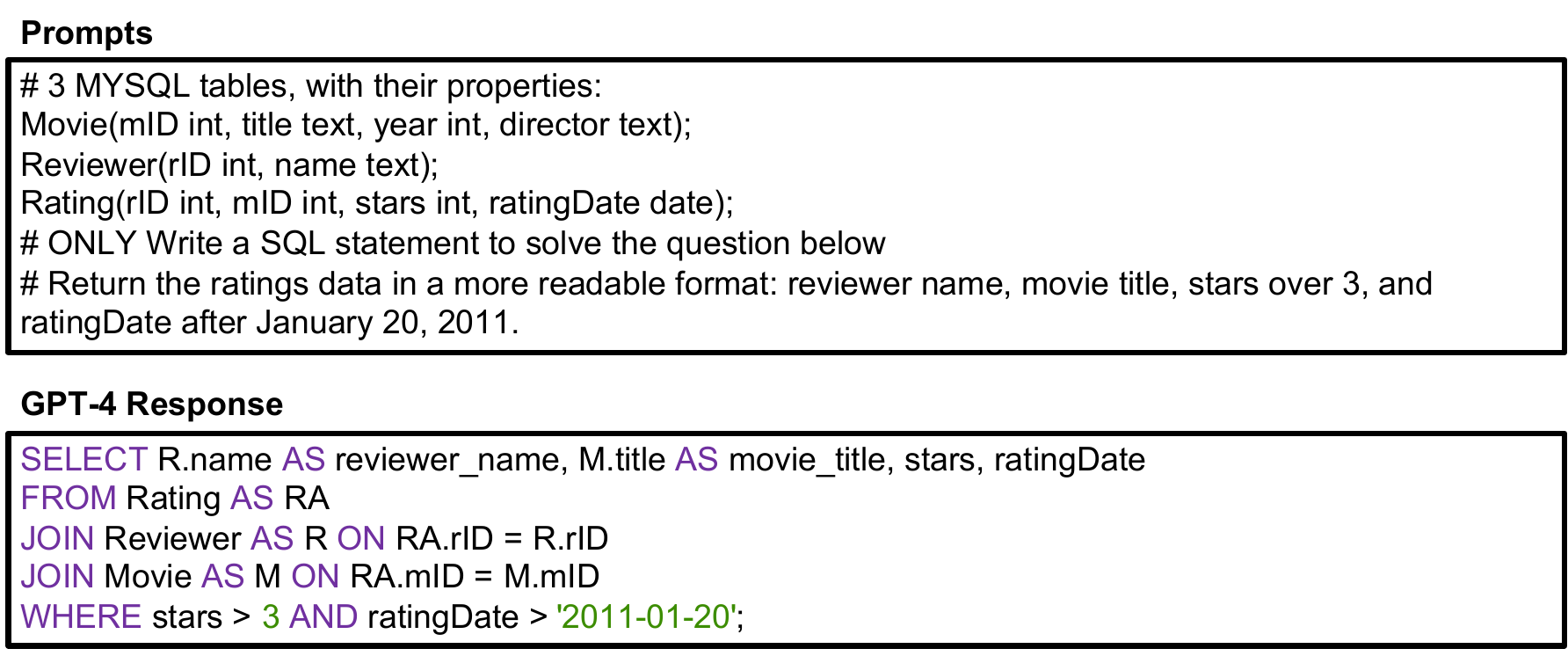}
    \caption{An Example of the NL2SQL as Tool-Uses}
    \label{fig:text2sql}
\end{figure*}

In Fig.~\ref{fig:ai-agent}, we present three commonly-adopted tool-uses in a LangChain, including databases, knowledge graphs, and the pre-trained model zoo. Specifically, the use of databases and knowledge graphs aims at providing the agent exact information and data for context modeling and reasoning. Actually, LLMs, such as ChatGPT and GPT-4, have been extensively pre-trained using an ultra-large amount of Internet texts (e.g., GPT-3 was trained using texts of 570 GB and 300 billion words), they have demonstrated their knowledge and comprehensive understanding of various domains available on the Internet. However, LLMs still make crucial mistakes on the essential facts when generating contents, due to hallucinations of LLMs in natural language generation~\cite{ji2023survey}. The use of databases and knowledge graphs can help the agent build-up the capacities of data-driven context modeling and rule-based context reasoning, which are compatible with the existing ontology-based solutions for the similar purposes~\cite{wang2004ontology}. Nowadays, a LangChain can easily integrate natural language interfaces with SQL and SPARQL for query and manipulation~\cite{pourreza2023din,meyer2023llm}. Fig.~\ref{fig:text2sql} illustrates an example of prompts and GPT-4 responses for SQL query generation, namely NL2SQL (Natural Language to SQL), to search movie reviews with rating over 3 stars from three MySQL tables. It is obvious that GPT-4 is capable of performing sophisticated ``cross-table'' reasoning for advanced database-uses, even without incorporating demonstrations in prompts.

%In addition to enable ubiquitous tool-uses, code generation and execution.

Furthermore, the use of pre-trained foundation models can extend the perception and AI capacity of a \TheName{} system, together with the LLM. For example, given an image as the source of vision for context modeling, one can pickup the pre-trained ViT or ResNet from the model zoo to extract a feature vector from the image, and then leverage the feature vector as the tokens of a continuous prompt for visual-language prompting with the LLM~\cite{tsimpoukelli2021multimodal,sollami2021multimodal,alayrac2022flamingo}. In this case, the LLM can read contents in the image and perform context reasoning accordingly. More recently, with the help of several pre-trained models, an generalist agent, namely GATO, has been proposed to work ``as a multi-modal, multi-task, multi-embodiment generalist policy'' beyond the realm of text outputs~\cite{reed2022generalist}. In conclusion, the fusion of pre-trained models and continuous prompts (introduced in Section II.B), which leverages the outputs of other pre-trained models as the input tokens of LLMs, provide us feasibility to largely extend the perception, reasoning, and even action capacity of a \TheName{} system via prompt-based tuning with low costs.

\subsection{Discussion on the \TheName{} Architecture}
To build-up \TheName{} systems, a range of LLM-empowered AutoAgents such as HuggingGPT~\cite{shen2023hugginggpt}, AutoGPT~\footnote{AutoGPT - \url{https://github.com/Significant-Gravitas/Auto-GPT}}, GPT-Engineer~\footnote{GPT-Engineer - \url{https://github.com/AntonOsika/gpt-engineer}}, and BabyAGI~\footnote{BabyAGI - \url{https://github.com/yoheinakajima/babyagi}} have been designed to proficiently interact with users and environments while offering effective solutions to complex problems. The overall goal of these systemms are to  actualize ``Embodied Artificial Intelligence'' (Embodied AI)~\cite{franklin1996agent,pfeifer2004embodied}, which broadens the proficiency of large language models and other foundational models to interact with environments via physical embodiment~\cite{chen2023next}.

To better understand practices with the \TheName{} architectural design and the LangChain-based integrative implementation of am AutoAgent, especially the prompts design and possible responses from LLMs, we present two showcases in following sections.

%On top of prompt techniques of LLMs, LangChain~\cite{langchain2023} has been proposed to ``chain'' together different components, including natural language user interfaces (NLI), varying foundation models, long-term memory, external knowledge bases, prompt templates, and interfaces of tool-uses, for building-up innovative LLM-powered autonomous applications~\cite{bubeck2023sparks}. 

\section{Demonstrating Showcases of \TheName{}}
In this section, we present two demonstrating showcases for building-up \TheName{} systems. While the first showcase demonstrates the prompts and responses for action planning, the second one demonstrates the prompts and responses to perform multi-level task decomposition to adapt contexts.

\begin{figure}%[btp]
    \centering
    \includegraphics[width=0.5\textwidth]{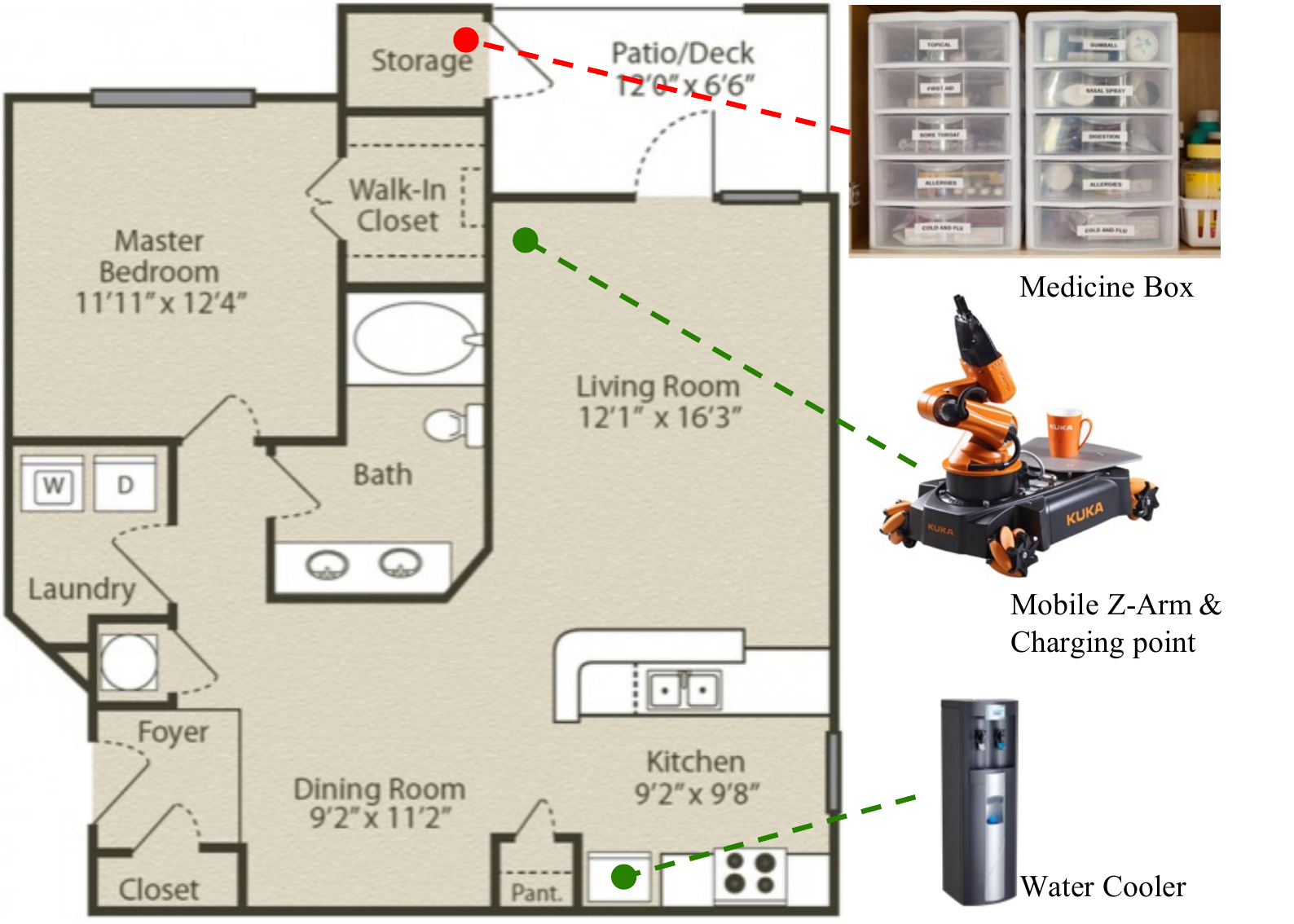}
    \caption{The layout of the apartment and settings of the task}
    \label{fig:room-layout}
\end{figure}

\subsection{Mobile Z-Arm for Assisted Living in an Apartment}
In this showcase, we demonstrate the way that, upon a user's request, GPT-4 generates the action plan and controls a mobile z-arm for medication purpose via the LangChain deployed at a smart speaker.

\subsubsection{System Settings} Fig.~\ref{fig:room-layout} shows a \TheName{} system that operates a mobile z-arm for assisted living in an apartment consisting of multiple rooms, including a living room, a bedroom, a kitchen, a bathroom, and a storeroom. The kitchen is equipped with a water cooler, while the storeroom houses a medicine box. It takes the mobile z-arm approximately 1 to 2 minutes to travel between each room, and the mobile z-arm must return to the charging port located in the living room once the task is complete. In this showcase, the user sends a simple request to a smart speaker (e.g., an Amazon Echo) as follow. 
\begin{itemize}
    \item[] \emph{``I need to take two pills of aspirin with water at 10pm in the living room''}
\end{itemize}
The smart speaker forwards the request to GPT-4 within a prompt and GPT-4 responds with an action plan. Further, the smart speaker follows the action plan and controls the mobile z-arm to finish the task.

\begin{figure*}%[btp]
    \centering
    \includegraphics[width=0.9\textwidth]{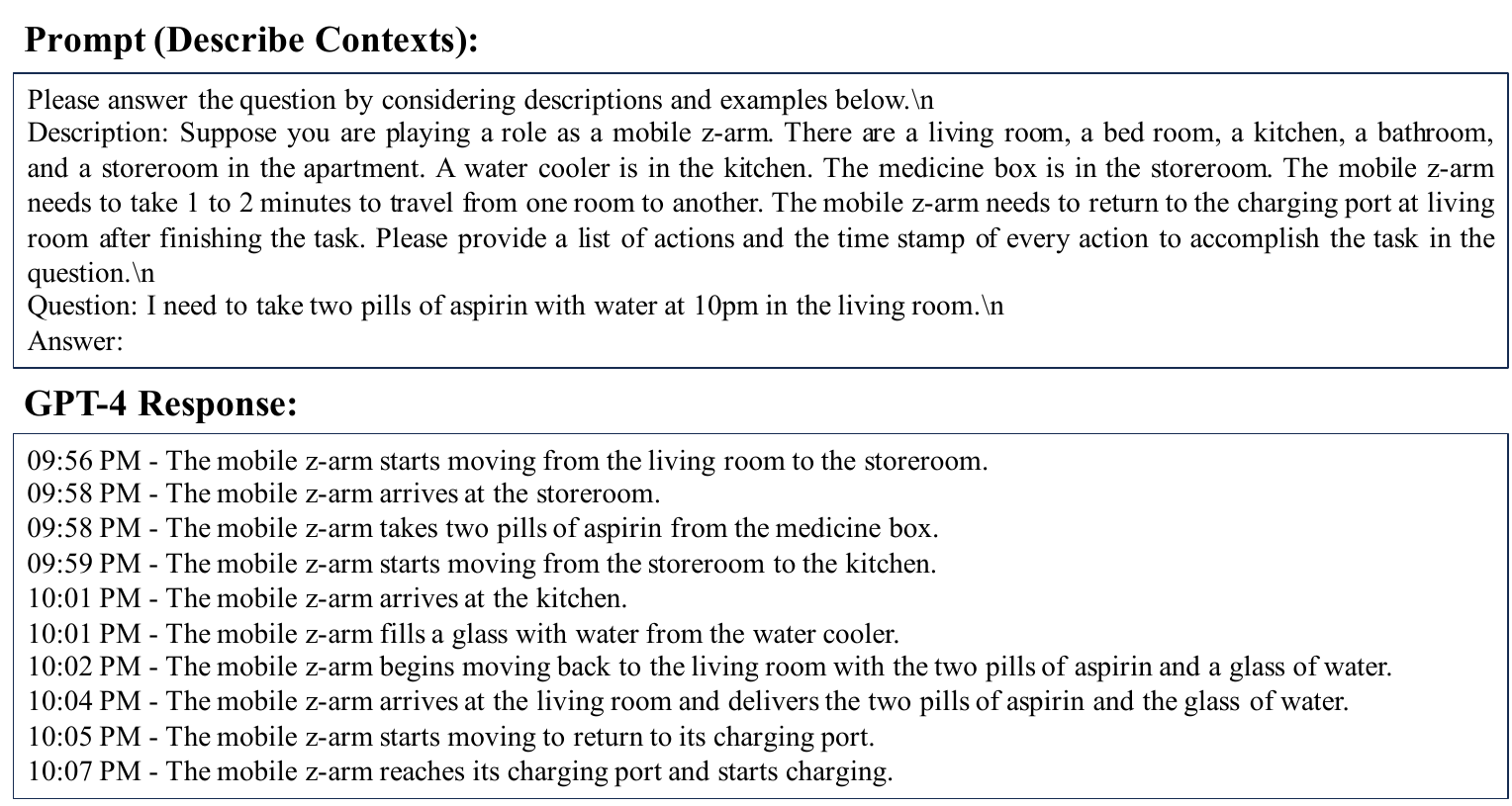}
    \caption{Zero-shot Prompting for Action Plan Generation to Control a Mobile Z-Arm for Medication Purpose}
    \label{fig:mobile-zarm-1}
\end{figure*}

\subsubsection{\TheName{} System Designs} We follow a simple design of \TheName{} system, which assumes to integrate the AutoAgent on smart speakers, water cooler, mobile z-arm, and other appliances via a LangChain. In the \TheName{} system, the LangChain leverages a \emph{Conversational Agent} listed in Algorithm~\ref{alg:ai-agent} to handle the user's request. Given the user's request, the agent first categorizes the type of request (i.e., ``take medicine''), then matches the template of prompt from a repository of prompt templates (for varying types of requests including ``take medicine'' and so on), subject to the request type. Later, the agent fills the texts of user's request and context information into the template to form the prompt. The agent calls the LLM by forwarding the prompt, and receives the response from the LLM to obtain the action plan. Finally, the agent follows actions in the plan to fulfill the user's request. For the sake of simplicity, the LangChain for this system could be ``stateless'' without incorporating the \emph{Conversational Memory} component.

\subsubsection{Prompts and Responses for Action Planning} 
\begin{figure*}%[btp]
    \centering
    \includegraphics[width=0.9\textwidth]{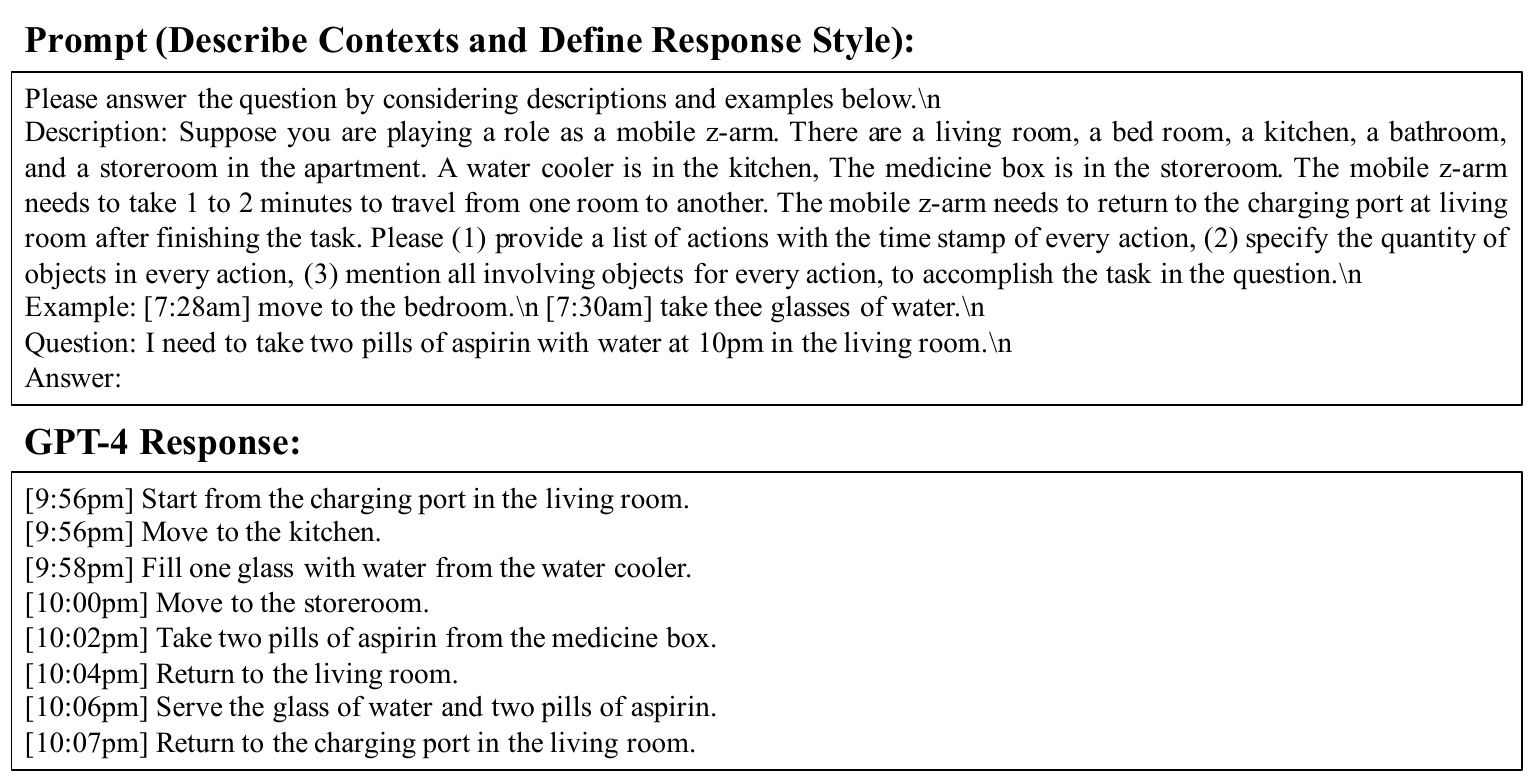}
    \caption{Few-shot Prompting for Action Plan Generation to Control a Mobile Z-Arm for Medication Purpose}
    \label{fig:mobile-zarm-2}
\end{figure*}

Fig.~\ref{fig:mobile-zarm-1} and Fig.~\ref{fig:mobile-zarm-2} present two sets of prompts and responses for making action plans. Fig.~\ref{fig:mobile-zarm-1} demonstrates a \emph{``zero-shot'' prompt} that includes a section of description to model the context of mobile z-arm operations and a question from the user's request. In the description section, the prompt first turns on GPT-4's role-playing mode, then describes the rooms and distribution of facilities in the apartment for action planning. With the prompt, GPT-4 respond the question with a human-readable action plan. In such plan, Z-arm moves from living room to storeroom at 9:56 PM; takes aspirin at 9:58 PM in storeroom; fills glass with water at 10:01 PM in kitchen; delivers aspirin and water to living room at 10:04 PM; returns to charging port at 10:05 PM and starts charging at 10:07 PM. It looks that such action plan can fulfill the request. However, the way GPT-4 formats the response might not be machine-readable.

Fig.~\ref{fig:mobile-zarm-2} illustrates a \emph{``few-shot'' prompt}, which in addition to \emph{the ``zero-shot'' prompt} adds one additional section of examples for expected responses. In this way, the prompt could define a machine-readable format of an action plan, while GPT-4 follows the example to respond in such pre-defined format automatically. Note that, it might be interesting to see the GPT-4 responses to these two prompts (left and right sides) are slightly different but both are correct in terms of actions. Especially, the responses become concise (with fewer actions) when introducing the section of examples in the prompt. Actually, the responses at the right side of Fig.~\ref{fig:mobile-zarm-1} (without any example in the prompt) tend to list all events that may happen with the mobile z-arm during the task, while responses at the right side (with examples in the prompt) only include the actions the mobile z-arm should act at every time to accomplish the task.

\subsection{Context-aware Personalized Trip Planner and Itinerary Scheduler}
In this showcase, we demonstrate yet another application of \TheName{}--the context-aware personalized trip planner~\cite{yang2013fine,yang2014modeling}, where LLM-powered AutoAgent has been used to recommend tourist attractions and schedule the itinerary based on users' past travelling records and the context of visits.

\subsubsection{System Settings} In this scenario, we assume the user has travelled a number of places and remained digital footprints on the user's past trips. For a trip to Paris, the user aims to leverage a \TheName{} system to plan the trip, schedule the itinerary, and act reservations accordingly. To enable context-awareness and personalization in this scenario,  we assume a \TheName{} system incorporate following two types of digital footprints as inputs.
\begin{itemize}
    \item \emph{Past travel records} include information on the tourist attractions, restaurants, theatres \& shows, and other related locations that the user has visited or engaged in the user's past journey. For example, the user visited Rome and Barcelona recently. The attractions that user visited include SS. Lazio (Stadio Olimpico), Trionfale, and the Pantheon for Rome Italy; as well as La Boqueria, and FC Barcelona (Spotify Camp Nou) for Barcelona, Catalonia. This records may indicates the user's preferences in historical sites, soccer, and local markets.
    
    \item \emph{Constraints of the itinerary} include information, such as the user's arrival or departure flights, hotels to stay, events to attend, and other temporal or geographical restrictions,  which constrain the current trip planning and scheduling in a mandatory manner. For example, the user will first arrive Paris by train at Gare de Saint-Cloud (17:10, 2023-07-23), then attend two academic events -- a research talk and a banquet -- in 2023-07-24, and finally depart from CDG Airport (13:00, 2023-07-25).
\end{itemize}
We further assume all above information could have been represented as texts, while \TheName{} leverages a LangChain and GPT-4 to achieve the goal.

\begin{figure*}
    \centering
    \includegraphics[width=\textwidth]{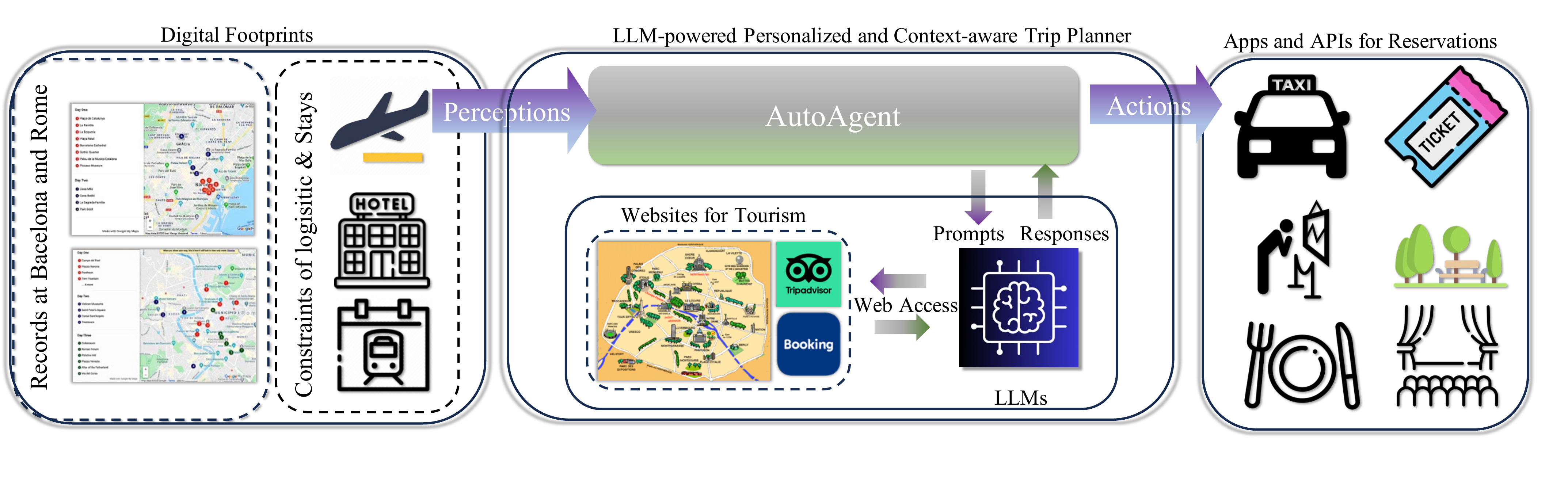}
    \caption{Design of a \TheName{} System for Context-aware Personalized Trip Planning, Itinerary scheduling, and Reservations}
    \label{fig:trip-planner}
\end{figure*}

\subsubsection{\TheName{} System Design}~Fig.~\ref{fig:trip-planner} illustrates the architectural design of the \TheName{} system for context-aware personalized trip planning, itinerary scheduling, and reservations. This system leverages an AutoAgent to form the user's past travel records and constraints of the itinerary into prompts and forward these prompts to LLMs for responses. The LLM is with web accesses and can use the contents on travel websites (e.g., Foursquare, TripAdvisor, or Meituan) to respond the AutoAgent a draft of trip plan and itinerary schedules. The user can further discuss with the AutoAgent on the plan and itinerary, for revision and updates. Once the plan and itinerary are fixed, the AutoAgent would inquiry the LLM for generating an action plan to do the reservations. Finally, the AutoAgent follows the action plan and call third-party APIs for reservations, including hotels, restaurants, and electronic ticketing of attractions.

\begin{figure*}[h]
    \centering
    \includegraphics[width=0.75\textwidth]{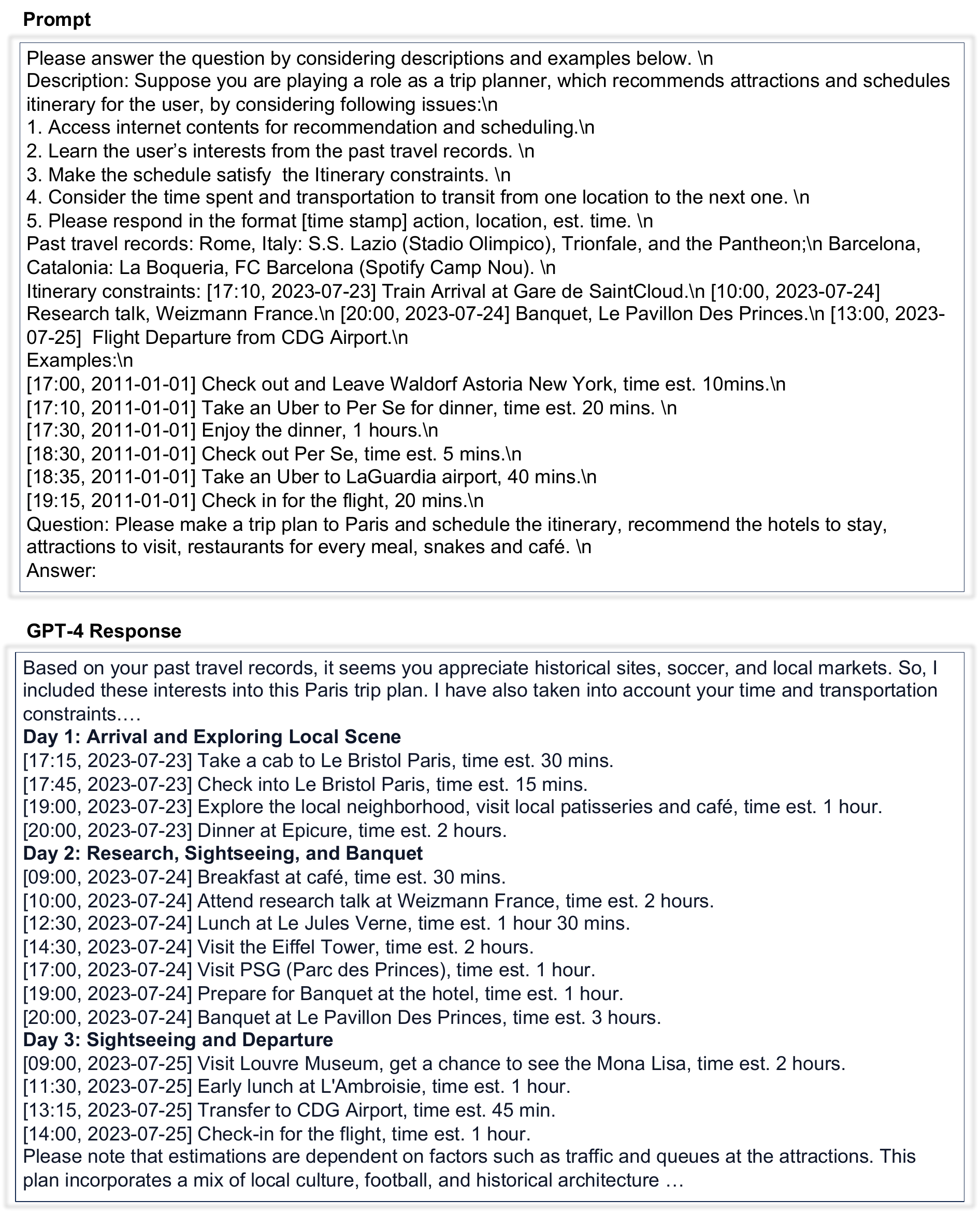}
    \caption{Generating a Trip Plan with Itinerary Schedules using Prompts and GPT-4}
    \label{fig:trip-plan-prompt}
\end{figure*}

\subsubsection{Prompts and Responses for Trip Planning and Itinerary Scheduling} Fig.~\ref{fig:trip-plan-prompt} presents the prompts and GPT-4 responses for the trip planning and itinerary scheduling. From a Personalization Perspective, this plan takes into account the user's interests learned from previous travel experiences, such as an affinity for soccer (S.S. Lazio in Rome and FC Barcelona in Barcelona), historical sites (the Pantheon and Trionfale in Rome), and local markets (La Boqueria in Barcelona). In addition, from a itinerary constraints perspective, the schedule allows for a 30-minute breakfast on 2023-07-24 before the research talk. It gives the user ample time to reach Weizmann Institute France on time. The itinerary takes into account the Banquet at 20:00 on 2023-07-24, while suggesting the user return to the hotel by 19:00 for freshening up. Regarding the flight departure, the proposed itinerary suggests a sufficiently early checkout, allowing time for potential traffic and check-in requirements at the airport.

\subsubsection{Action Implementation}
\begin{figure*}
    \centering
    \includegraphics[width=0.78\textwidth]{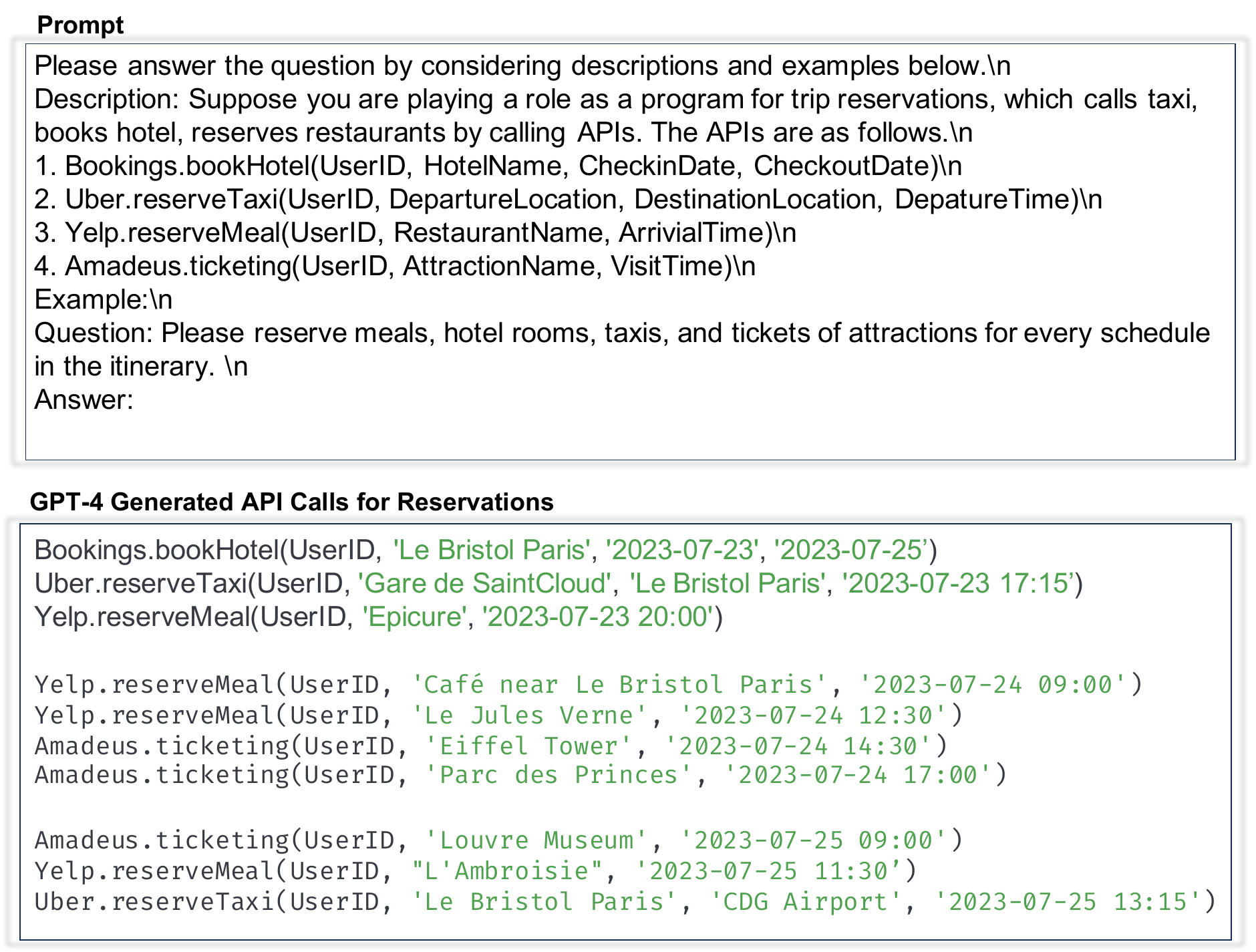}
    \caption{Code Generation Example of API Calls for Reservations using Prompts and GPT-4}
    \label{fig:api-calls-prompt}\vspace{-3mm}
\end{figure*}

To carry out the trip plan, the AutoAgent leverages multi-round conversations and appends an additional prompt after the response shown in Fig.~\ref{fig:trip-plan-prompt}, so as to make the LLM generate codes for API calls. Specifically, Fig.~\ref{fig:api-calls-prompt} shows an example of code generation, where (1) the specifications to the API calls are given in the prompts and (2) the GPT-4 responds the generated codes to call ``hypothetical'' APIs for booking rooms of hotel, reserving taxi for outgoings, reserving restaurants for meals, and electronic ticketing for attractions. We understand the generated codes might not be able to work directly. However, they demonstrate the feasibility of using prompts to generate codes for reservation purposes.

\subsection{Summary and Discussions}
In the above showcases, we clearly demonstrate using natural language and LLMs for context modeling and reasoning, while the result of context reasoning is an action plan generated by LLMs. Later, the agent follows the action plan to fulfil the user's request. Furthermore, the LLM, such as GPT-4, can quickly learn to respond in a way we desire through few-shot prompting, which only needs to include some carefully written examples in the prompt.

However, this system also suffers critical performance issues. For the identical prompt, GPT-4 might respond in different and non-deterministic ways. Actually, GPT-4 is designed to generate dynamic and creative responses, not deterministic ones. Therefore, even with the same prompt, multiple executions could result in different outputs. For example, in the showcase of \emph{mobile z-arm} (i.e., Figs.~\ref{fig:mobile-zarm-1} and~\ref{fig:mobile-zarm-2}), GPT-4 has generated an action plan to fist visit kitchen for water and then move to the storeroom for fetching the medicine. However, in another run of GPT-4, we found it has made a plan to first visit the storeroom for aspirin first and then move to the kitchen for water. Fortunately, the order of these two actions would not affect the overall results to fulfill the request here. Furthermore, the format of response could be varying in different executions. Even though we could use examples in few-shot prompting to define the format of outputs, the generation of action plan might be in different. For example, GPT-4 has described an action as ``[9:56pm] Move to the kitchen'' in Fig.~\ref{fig:mobile-zarm-2}, which has been also described as ``Move from the living room to the kitchen'' in some other runs. Similarly, in the showcase of \emph{trip planning and scheduling} (i.e., Fig.~\ref{fig:trip-plan-prompt}), the recommended attractions might be varying in different runs of GPT-4. At times, the suggested itinerary highlights only the most popular tourist destinations, like the Eiffel Tower, Mus{\'e}e d'Orsay, and Louvre Museum, while overlooking ``lesser-known'' attractions that might be of particular interest to the user, such as the PSG Football Club (Parc des Princes).

To improve the performance, here are some takeaways. First of all, one should  adjust and set the ``temperature'' of GPT-4 content generation to a relatively low level. For LLMs of GPT-alike, a temperature parameter is used to control the randomness of content generation. The temperature parameter essentially adjusts the generated output from purely deterministic (with temperature close to 0) to highly creative or diversified (with temperature close to 1). Note that, an extremely low temperature making LLMs lack of creativity (e.g., capacities of planning or imagination) should be avoid. Furthermore, the LangChain should equip with syntax analysis tools or text parsers for actions, to ensure the correctness in following the plan. Finally, one could train and deploy their own LLMs, using well-annotated requests and action plans as training data for SFT.

\section{Discussion and Future Works}
In this section, we discussion the limitations of \TheName{} and potential future works. We first compare the strength and weakness of \TheName{} with existing solutions. Later, we discuss the performance affecting factors of \TheName{} and shed the light to the future research directions in following subsections.

\subsection{Benefits and Limitations of \TheName{}}
Table~\ref{tab:context-aware} presents the benefits and limitations of the three approaches (\TheName{}, Widget-centric~\cite{dey2001conceptual} and Ontology-based~\cite{wang2004ontology} approaches) for context-aware computing.

%In terms of strength of \TheName{}, the natural language used in \TheName{} is flexible and easy to use for communication with users, which is a key advantage over the widget-centric approach's dependence on users' presence at a mobile terminal for interaction. Further, \TheName{} offers flexibility, extensibility, and adaptability for varied context reasoning and new observations. This is in contrast to the ontology-based and widget-centric approaches which are not adaptive to new samples or are limited in extension capabilities.

\begin{table}[h]
\centering
\caption{Strengths and Weaknesses of Widget-based, Ontology-based, and LLM-driven Context-aware Computing}\vspace{-3mm}
\begin{tabular}{p{0.1\textwidth}|p{0.27\textwidth}|p{0.27\textwidth}|p{0.27\textwidth}}
\hline
\textbf{Approach} & \textbf{Interfacing \& Prototyping} & \textbf{Context Reasoning} & \textbf{Context Modeling} \\ \hline
Widget-centric~\cite{dey2001conceptual} & Widgets offer graphical interfaces for user interaction but often require user's presence at a terminal. They can efficiently interface with a few sensors \& actuators by limited accessibility. & Rule-based reasoning could provide exact inference results, but is not extensible due to the lack of standard data formats rules. Traditional machine learning-based solutions are task-specific and require redesigning or retraining models to accommodate new samples or tasks. & Tuples are efficient in data storage, retrieval, and communication, but rely on pre-defined schema and lack of open and standard data formats for exchange and integration.  \\ \hline

Ontology-based~\cite{wang2004ontology} & Semantic Web Services are extensible solutions to integrate various sensors \& actuators upon open and standard data formats, such as OWL, but also require user's presence at a terminal for interactions. & Ontology-based~reasoning can provide exact results but limited on description logic~\cite{baader2003description}. Rules  for reasoning might be extensible with SWRL~\cite{o2005supporting} but not adaptive to new samples. & Ontology/OWL is an open and standard data formats allowing data integration and manipulation, but also relying on pre-defined schema.\\ \hline
\TheName{} &  Natural Language is flexible and easy to use for communication with users, but might be ambiguous and inefficient. & AutoAgents on LLMs offer flexibility, extensibility and adaptability for varied context reasoning and new samples. However, the reasoning results can be ambiguous and hard to interpret.  & Texts are flexible to represent contexts of different types, but might be ambiguous and inefficient. \\ \hline
\end{tabular}
\label{tab:context-aware}\vspace{-3mm}
\end{table}

Concerning the merits of \TheName{}, the use of natural language for user interactions stands out as both adaptable and user-friendly, offering a significant benefit over the widget-centric approach which requires the presence of users at a mobile terminal for interaction. Moreover, \TheName{} radiates versatility, extendibility, and adaptability when handling a variety of contextual reasoning and newly arising observations. This manifestly sets TheName apart from the ontology-based and widget-centric approaches, both of which display restrictive flexibility when dealing with new samples and have limited scalability.

%In terms of weakness of \TheName{},  the natural language, however, can often be ambiguous and inefficient, especially compared to the efficiency and precision of the widgets or Semantic Web Services in the Widget-Centric and Ontology-based approaches, respectively. Furthermore, the context reasoning results of \TheName{} can be ambiguous and hard to interpret. This is in contrast to the exact inference results offered by pre-defined rules in ontology-based reasoning. Finally, the use of texts to represent contexts can also be ambiguous and inefficient in comparison to tuple usage in the widget-centric approach or the open and standard data formats of the ontology-based approach.  

Regarding the limitations of \TheName{}, its usage of natural language can, at times, introduce ambiguity and inefficiency, particularly when compared to the precision delivered by widgets in the widget-centric approach and semantic web services in the ontology-based approach. Furthermore, the reasoning outcomes provided by \TheName{} can be multifaceted and challenging to interpret, standing in contrast to the exact inference results guaranteed by the pre-defined rules in ontology-based reasoning. Lastly, the use of texts to represent contexts may also give rise to ambiguity and inefficiency when viewed against the structured tuples used in the widget-centric approach or the open and standard data formats employed by the ontology-based approach.

%In general, widget-based approaches are effective with dedicated programs communicating with sensors \& actuators but can be less flexible due to the constraints of pre-defined schemes for context data and rules. Ontology-based methods offer extensible integration due to open data formats but have limited adaptability to new samples. LLM-driven strategies are flexible and adaptive but may present inefficiencies due to potential ambiguity of natural languages.

%\subsection{Limitations and Future Directions}

\subsection{Performance Affecting Factors}
Actually, there are a few factors that may affect the performance of context-aware computing. Here we discuss several of them from the perspectives of LLMs as follows.
\begin{itemize}
    \item \emph{Token Limits in Input and Output Texts of LLMs:} The number of tokens in input and output sequences significantly impacts LLM performance~\cite{bubeck2023sparks,openai2023gpt4}. Shorter input sequences may lack necessary information, leading to reduced accuracy and coherence, while longer inputs can improve performance but may cause computational inefficiency. Similarly, shorter output sequences can result in incomplete or less informative responses, whereas longer outputs provide more detailed responses but risk generating irrelevant or repetitive text. Therefore, optimizing token limits in both input and output texts is essential for balancing performance, accuracy, and coherence in LLM-generated contents~\cite{bubeck2023sparks}.

    \item \emph{Long-term and Hierarchical Action Planing:} LLM can break down complex tasks into manageable sub-plans using simple prompts or human inputs, which is essential for context reasoning. In addition to the use of external planners~\cite{liu2023llm+}, Chain of Thoughts (CoT) and Tree of Thoughts (ToT) techniques~\cite{wang2022self,yao2023tree} have been proposed to enable multi-level decomposition by instructing the model to think step by step and explore multiple reasoning possibilities. Furthermore, LangChain integrates LLM with CoT or ToT to create hierarchical plans spanning extended periods and decompose tasks into subtasks at varying levels of abstraction~\cite{langchain2023}. This approach employs natural language prompts for both plan formation and action execution. A comprehensive online tutorial could be found from the weblog~\cite{lilian2023LLM}.

    \item \emph{Trustworthiness and Interpretability of LLMs:} The trustworthiness and interpretability of LLMs greatly impact the performance of LangChain for action planning and execution. Trustworthy LLMs generate reliable action plans with high-quality, well-structured and hallucination-free contents~\cite{bang2023multitask}, while high interpretability allows users to understand the ways of LLM-based reasoning, facilitating issue identification and process refinement~\cite{li2022interpretable}. Ensuring that LLMs are both trustworthy and interpretable will lead to more effective, safe, and robust action planning and execution using LangChain~\cite{langchain2023,lilian2023LLM}.

\end{itemize}
Table~\ref{tab:affect} provides an overview on the performance affecting factors, their technical challenges, and potential solutions for \TheName{}.

\begin{table}[ht]
\centering
\caption{Summary of factors affecting LLMs in context-aware computing.}\vspace{-3mm}
\begin{tabular}{p{0.15\textwidth}|p{0.32\textwidth}|p{0.4\textwidth}}
\hline
\textbf{Topic} & \textbf{Challenges} & \textbf{Potential Solutions} \\
\hline
Token Limits & Short sequences might be insufficient or vague, while long ones would cause inefficiencies and irrelevant outputs. & Scaling LLMs to handle extremely long sequences~\cite{xue2023repeat}, Leveraging long-term memory~\cite{LTM1,liang2023unleashing,zhong2023memorybank}. \\
\hline
Action Planning &  Effectively Break down complex tasks into manageable sub-plans while maintaining context reasoning. & External Planner~\cite{liu2023llm+} as tool-uses, Advanced LangChain techniques~\cite{lilian2023LLM}, including Chain of Thoughts (CoT)~\cite{wang2022self} and Tree of Thoughts (ToT)~\cite{yao2023tree}. \\
\hline
Trustworthiness  & Generating reliable action plans with hallucination-free, high-quality, and well-structured contents. & Automatic hallucination checking~\cite{manakul2023selfcheckgpt} or fact-checking~\cite{lee2020language},  Human alignments via intensive SFT or RLHF~\cite{ouyang2022training}, Explainable (XAI)~\cite{li2022interpretable} for LLMs' behavior understanding. \\
\hline
\end{tabular}
\label{tab:affect}
\vspace{-3mm}
\end{table}

\subsection{Future Research Directions}
In addition, the potential future directions along the line of research in this field are as follows.
\begin{itemize}
    \item \emph{Pre-trained Foundation Models:} In this work, we mainly discuss the use of large language models (LLMs) to enable context-awareness with texts, prompts, LangChain, and external devices. Though it is possible to model contexts using texts and perform reasoning accordingly, there still need incorporate additional modalities, such as vision, acoustic, and haptic to enrich the context characterization~\cite{huang2023visual,mu2023embodiedgpt,liu2023visual}.  In addition, the capacities of LLMs could be further improved, with advanced transformer architectures (e.g., decoder-only versus encoder-decoder versus GLM)~\cite{xue2021mt5,zeng2022glm,anil2023palm}, and additional pre-training datasets, strategies, and tasks~\cite{radford2018improving}.

    \item \emph{Domain-specific Tuning:} Given a pre-trained foundation model, there frequently needs to further train the model to adapt specific application domains, such as healthcare~\cite{yunxiang2023chatdoctor}. According to~\cite{openai2023gpt4,bubeck2023sparks}, domain-specific tuning strategies can be divided into several training phases that follow the pre-training phase. These phases, in order, are post pre-training (PPT), supervised fine-tuning (SFT), and prompt tuning (P-tuning). While PPT involves further training on domain-specific data via the autoregressive tasks~\cite{radford2018improving} after initial pre-training, while SFT focuses on using labeled data (e.g., question-answer pairs) to enhance task-specific performance. Furthermore, P-tuning, on the other hand, usually inserts additional learnable layers into the LLM and tunes continuous prompts with the model remaining frozen, aiming to reduce the memory usage during training~\cite{liu2022p}. P-tuning can even help LLMs to adapt new domains in different modalities, such as images~\cite{liu2023visual}.

    \item \emph{Task Decomposition with CoT and ToT:} Task decomposition is a crucial aspect of LLM-powered autonomous agents. It involves breaking down complex tasks into smaller, more manageable subtasks. This process aids in interpreting the model's thinking process and improves the model's performance on complex tasks. Chain of thought (CoT) and Tree of Thoughts (ToT) are two common techniques used for task decomposition. CoT transforms big tasks into multiple manageable tasks and provides insights into the model's interpretation. ToT goes a step further by exploring multiple reasoning possibilities at each step. Task decomposition can be achieved through various methods. LLMs can use simple prompting techniques, task-specific instructions, or even incorporate human inputs to decompose tasks effectively. This approach enhances the agent's ability to plan ahead, make informed decisions, and improve overall performance.

    \item \emph{Gateway of Physical Embodiment:} AutoAgents can interact with the world more effectively, perceive and sense the physical environment, and understand the context more comprehensively through tool-uses and physical embodiment. By utilizing tools and APIs, AutoAgents can expand their capabilities and problem-solving abilities, and enable real-time adaptability, empowering them to dynamically respond to changing circumstances. However, with rapid development of \TheName{}, there might need a gateway with a unified set of communication protocols to facilitate the interactions between AutoAgents and the physical embodiment of varying types.

    %\item \emph{Tokenization and General Agency:}
    
    %Afterall, these capabilities enhance the potential of autoagents to provide more comprehensive and effective assistance to users in a wide range of contexts.

\end{itemize}
Due to the page limits, we haven't include many other performance affecting factors or potential research directions, such as the scaling LLMs for emergence of abilities~\cite{bubeck2023sparks}, here.

\section{Conclusions}
LLMs have gained significant attention in recent years, with their natural language understanding capacities making it possible to model and infer contexts by natural language based context representation. In this work, we have explored the potential of LLMs in driving context-aware computing, a paradigm we refer to as LLM-driven Context-aware Computing (\TheName{}). Specifically, \TheName{} integrates users, LLMs, autonomous agents (AutoAgents) on ubiquitous devices, and sensors \& actuators in a LangChain, where LLMs can perform context modeling and reasoning without the need for model fine-tuning. By representing users' requests, sensor data, and actuator commands as text, the AutoAgent can model context and interact with LLMs for reasoning and generating action plans. With the action plan responded by LLMs, the AutoAgent follows the plan by executing actions inside to fulfill the user's request.

We provided two practical showcases to illustrate the effectiveness of the proposed \TheName{} architecture: (1) operating a mobile z-arm in an apartment for assisted living, and (2) planning a trip and scheduling the itinerary in a context-aware and personalized manner. These examples highlight the versatility and potential of \TheName{} in various applications. Moreover, we analyzed several factors that may influence the performance of \TheName{} and discussed future research directions. As LLMs continue to evolve and improve, we anticipate that their role in context-aware computing will become increasingly significant, paving the way for a wide range of innovative applications in healthcare, transportation, smart homes, retail, and beyond.

\bibliographystyle{unsrt}

\bibliography{main}

\end{document}